\documentclass{article}
\usepackage{arxiv}
\usepackage{tabularx} 
\usepackage{booktabs} 
\usepackage{caption} 
\usepackage{geometry} 
\usepackage[utf8]{inputenc} 
\usepackage[T1]{fontenc} 
\usepackage{hyperref} 
\usepackage{url} 
\usepackage{booktabs} 
\usepackage{amsfonts} 
\usepackage{nicefrac} 
\usepackage{microtype} 
\usepackage{graphicx}
\usepackage{natbib}
\usepackage{doi}
\usepackage{amsmath, amssymb, amsthm}
\usepackage{mathtools}
\usepackage{enumitem}
\usepackage{xcolor}
\usepackage{caption}
\usepackage{subcaption}
\usepackage{algorithm}
\usepackage{algpseudocode}
\usepackage{siunitx}
\usepackage{bm}
\usepackage{cleveref}
\usepackage{etoolbox}
\usepackage{listings}
\lstdefinelanguage{JavaScript}{
  keywords={typeof, new, true, false, catch, function, return, null, catch, switch, var, if, in, while, do, else, case, break},
  keywordstyle=\color{blue}\bfseries,
  ndkeywords={class, export, boolean, throw, implements, import, this},
  ndkeywordstyle=\color{darkgray}\bfseries,
  identifierstyle=\color{black},
  sensitive=false,
  comment=[l]{//},
  morecomment=[s]{/*}{*/},
  commentstyle=\color{purple}\ttfamily,
  stringstyle=\color{red}\ttfamily,
  morestring=[b]',
  morestring=[b]"
}

\hypersetup{
pdftitle={Autonomous Navigation of Cloud-Controlled Quadcopters in Confined Spaces Using Multi-Modal Perception and LLM-Driven High Semantic Reasoning},
pdfsubject={Robotics, Artificial Intelligence, Autonomous Navigation},
pdfauthor={Shoaib Ahmmad, Zubayer Ahmed Aditto, Md Mehrab Hossain, Noushin Yeasmin, Shorower Hossain},
pdfkeywords={Perception, Quadcopter, GPS-denied, YOLOv11, Depth Anything V2, Large Language Model, Vision Language Model, Sensor Fusion, High Semantic Reasoning},
}
\title{Autonomous Navigation of Cloud-Controlled Quadcopters in Confined Spaces Using Multi-Modal Perception and LLM-Driven High Semantic Reasoning}
\author{
  Shoaib Ahmmad \\
  Department of Mechanical Engineering \\
  Rajshahi University of Engineering and Technology \\
  Rajshahi, Bangladesh \\
  \texttt{shoaibniloy434545@gmail.com} \\
  \And
  Zubayer Ahmed Aditto \\
  Department of Industrial and Production Engineering \\
  Shahjalal University of Science and Technology \\
  Sylhet, Bangladesh \\
  \texttt{zubayerahmed0017@gmail.com} \\
  \And
  Md Mehrab Hossain \\
  Department of Industrial and Production Engineering \\
  Bangladesh University of Engineering and Technology \\
  Dhaka, Bangladesh \\
  \texttt{mehrabshuvo30@gmail.com} \\
  \And
  Noushin Yeasmin \\
  Department of Urban and Regional Planning \\
  Rajshahi University of Engineering and Technology \\
  Rajshahi, Bangladesh \\
  \texttt{noushinoishe12@gmail.com} \\
  \And
  Shorower Hossain \\
  Department of Computer Science Engineering \\
  United International University \\
  Dhaka, Bangladesh \\
  \texttt{shorower777@gmail.com} \\
}
\usepackage{epstopdf}
\begin{document}
\maketitle
\begin{abstract}
This paper introduces an advanced AI-driven perception system for autonomous quadcopter navigation in GPS-denied indoor environments. The proposed framework leverages cloud computing to offload computationally intensive tasks and incorporates a custom-designed printed circuit board (PCB) for efficient sensor data acquisition, enabling robust navigation in confined spaces. The system integrates YOLOv11 for object detection, Depth Anything V2 for monocular depth estimation, a PCB equipped with Time-of-Flight (ToF) sensors and an Inertial Measurement Unit (IMU), and a cloud-based Large Language Model (LLM) for context-aware decision-making. A virtual safety envelope, enforced by calibrated sensor offsets, ensures collision avoidance, while a multithreaded architecture achieves low-latency processing. Enhanced spatial awareness is facilitated by 3D bounding box estimation with Kalman filtering. Experimental results in an indoor testbed demonstrate strong performance, with object detection achieving a mean Average Precision (mAP50) of 0.6, depth estimation Mean Absolute Error (MAE) of 7.2 cm, only 16 safety envelope breaches across 42 trials over approximately 11 minutes, and end-to-end system latency below 1 second. This cloud-supported, high-intelligence framework serves as an auxiliary perception and navigation system, complementing state-of-the-art drone autonomy for GPS-denied confined spaces.
\end{abstract}
\crefname{figure}{Figure}{Figures}
\crefname{table}{Table}{Tables}
\crefname{section}{Section}{Sections}
\section{Introduction}
Imagine a drone navigating through the debris of a collapsed building to locate survivors, operating entirely without GPS signals. This scenario underscores the core challenge of autonomous indoor drone navigation: the absence of satellite-based localization, the unpredictability of dynamic obstacles (e.g., shifting objects or moving people), and the requirement for real-time computation in constrained and cluttered environments. These conditions demand intelligent, compact, and robust systems—especially critical for applications in search and rescue, infrastructure inspection, and environmental monitoring \citep{nahavandi2025}.
Indoor navigation differs significantly from outdoor operations, where GPS and expansive sensor suites are commonly employed. Instead, indoor UAV systems must operate with lightweight sensors and limited onboard processing, requiring seamless integration of perception, control, and decision-making. While advances in robotics, artificial intelligence (AI), and embedded sensing technologies have accelerated progress, significant gaps remain in the development of fully autonomous, context-aware systems \citep{chang2023, elmokadem2021}.
Existing approaches typically address these challenges in isolation. For instance, Simultaneous Localization and Mapping (SLAM) techniques \citep{munguia2025, zhao2025, jamaludin2025}, such as ORB-SLAM3 \citep{campos2021orbslam3} and MASt3R-SLAM \citep{murai2025mast3r}, provide accurate pose estimation and mapping. However, they primarily function at the geometric level and are incapable of high-level semantic reasoning or contextual understanding. These algorithms cannot interpret the meaning of objects, predict dynamic behaviors, or respond to high-level commands—capabilities essential for intelligent autonomy in dynamic indoor environments. Even recent deep visual SLAM methods like DROID-SLAM \citep{javier2024} or GAN-enhanced SLAMs \citep{chang2023gan, smith2024} are limited in semantic awareness and struggle under low-visibility conditions.
Similarly, components like reinforcement learning (RL) for control adaptability \citep{drone_racing2024, uwano2024}, computer vision for obstacle detection \citep{uav_cv_review2024, cfyolo2025}, and emerging systems using vision–language models (VLMs) and large language models (LLMs) \citep{llmuav2024, llmland2025} have shown great promise. YOLO-based frameworks offer fast object detection \citep{cfyolo2025, li2024yolo, wang2024yolo, perot2024}, monocular depth networks such as Depth Anything V2 yield rich spatial understanding \citep{yang2024depthanything}, and LLMs enhance semantic reasoning and contextual awareness \citep{cai2025llmland, tian2025}. Additionally, works like \citep{kim2023, jonas2025} demonstrate that monocular depth estimation is becoming a cornerstone of geometry-aware UAV navigation, yet still suffer from limited integration into higher-level semantic reasoning systems.
However, these methods have largely been used in isolation, without forming a unified, real-time pipeline. This fragmentation prevents fully leveraging the synergy of deep vision models and reasoning modules, as LLMs remain underutilized due to their computational demands \citep{brown2020, liu2025ellmer, chen2024leviosa}. Recent VLM-based UAV frameworks \citep{liu2024} emphasize the need for unified benchmarks and platforms that incorporate reasoning alongside vision, but implementations remain fragmented and experimental.
This siloed approach leads to a fragmented perception system incapable of robust decision-making in unstructured spaces. SLAMs lack semantic awareness; detectors lack geometric context; LLMs demand computational resources beyond what most UAVs can handle; and combined, they are rarely optimized for concurrency and real-time inference. Furthermore, many solutions rely on LiDAR or multicamera rigs, increasing hardware costs and limiting scalability \citep{munguia2025}. Research such as \citep{chen2024rl, mnih2015, wang2021} suggests that scalable integration of RL-based control with vision-based perception and reasoning is key to closing this gap.
To overcome these limitations, this paper proposes a cloud-based, unified, multithreaded perception and decision-making framework for indoor drone navigation. The system integrates cutting-edge modules—including YOLOv11 for object detection \citep{cfyolo2025}, Depth Anything V2 for monocular depth estimation \citep{yang2024depthanything}, and LLM-based reasoning for semantic understanding \citep{cai2025llmland}—into a cohesive pipeline. A custom PCB interfacing six Time-of-Flight (ToF) sensors and an Inertial Measurement Unit (IMU) augments the visual feed, enabling precise spatial awareness. A multithreaded architecture processes these inputs in parallel, optimizing both latency and modularity. Computationally intensive tasks, such as LLM inference, are offloaded to cloud infrastructure to retain responsiveness.
This integrated system represents a decent leap forward. Unlike prior work that addresses SLAM, perception, or control in isolation, this research synthesizes all three domains into a lightweight, real-time, and scalable architecture capable of intelligent reasoning and adaptive autonomy. The drone can now perceive objects, estimate depth, interpret contextual semantics, and act upon this fused information in real-time.
This work introduces a novel AI-driven framework that bridges the gap between geometric navigation and semantic understanding. By leveraging the combined strengths of SLAM, deep learning, and LLMs in a real-time pipeline, it sets a new standard for intelligent UAV navigation in complex, GPS-denied indoor environments.
\section{Materials and Methods}
\subsection{System Architecture Overview}
\begin{figure}[htbp]
\centering
\includegraphics[width=\textwidth]{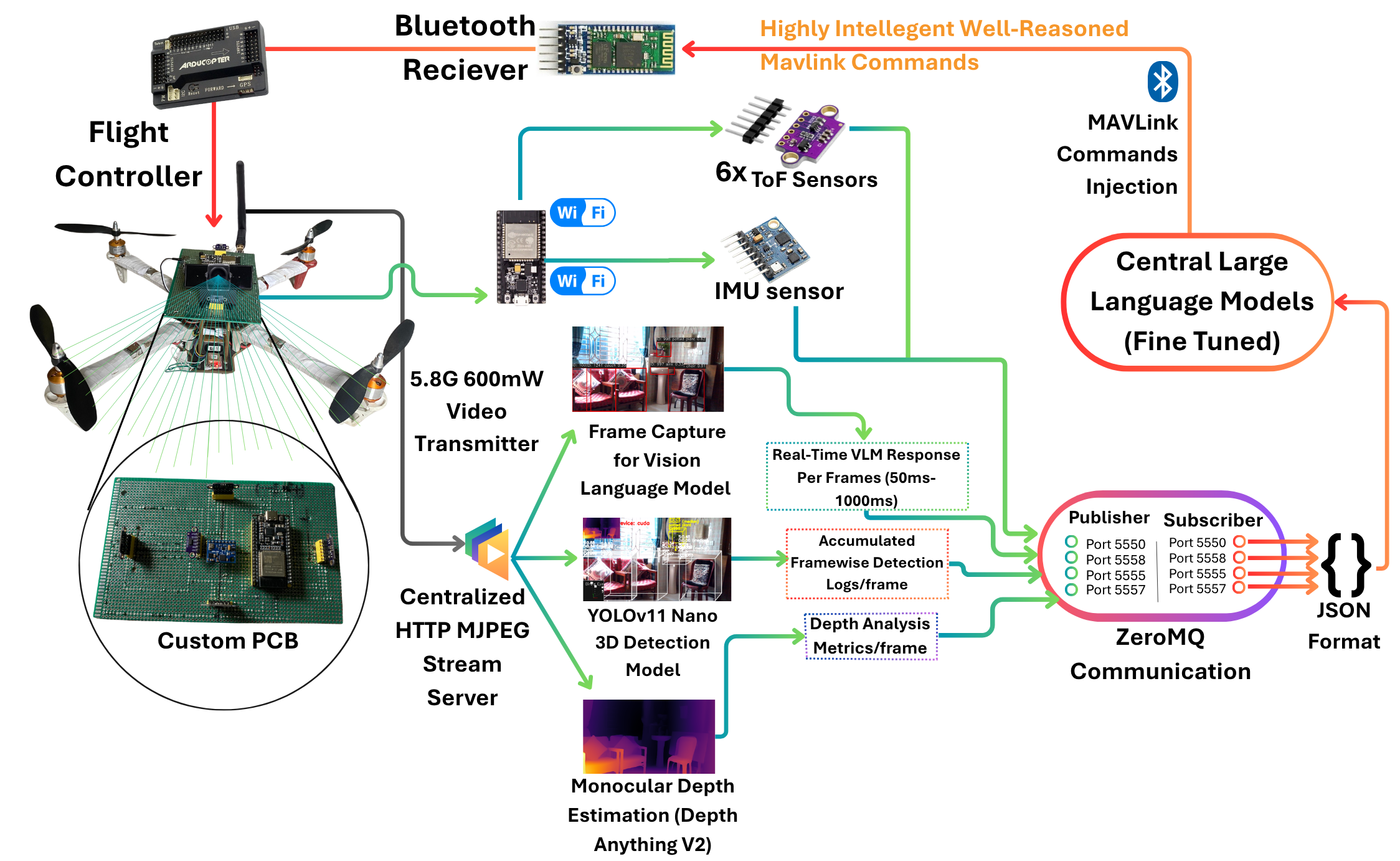}
\caption{Schematic Diagram of the System Architecture.\label{fig:system_architecture}}
\end{figure}
The drone's control system employs a real-time feedback loop, integrating data from an Inertial Measurement Unit (IMU) and six Time-of-Flight (ToF) sensors on a custom PCB. These sensors provide 3D environmental distance and orientation, sampling every 10 ms. Data are transmitted via Wi-Fi using an ESP32 sender, as shown in Figure~\ref{fig:system_architecture}, to a central Large Language Model (LLM). It also maintains a protective shield in confined spaces and allows smooth task transitions.
Concurrently, the drone's onboard camera streams footage to a cloud computing unit (in our case, a laptop). The YOLOv11 Nano model performs 3D object detection, with frames analyzed by a Local Vision Language Model (VLM) for semantic context. Monocular depth estimation, using Depth Anything V2, improves spatial awareness with a single lens camera.
Real-time data from YOLO, VLM, depth estimation, IMU, and ToF sensors are published via ZeroMQ (ZMQ). Raw inputs are JSON-formatted before being fed into the LLM. The LLM generates context-sensitive reasoning, sent as MavLink commands to the drone via Wi-Fi through a dedicated ESP32 receiver.
This hybrid system integrates multimodal sensory data, NLP, and advanced control to improve decision making, surpassing traditional autonomous navigation. By combining monocular vision, object detection, depth estimation, and semantic reasoning, this pipeline enables intelligent, context-aware drone navigation in dynamic environments.
\subsection{Camera Server}
We created a camera server to capture and broadcast video frames from the 5.8G OTG FPV Camera Combo 600mW CMOS 1200TVL Camera mounted on our drone at discrete time steps \( t_k = k \Delta t \), where \( k \in \mathbb{N} \) is the frame index and \( \Delta t = \frac{1}{30} \) seconds for 30 FPS. Each frame, denoted \( \mathbf{I}_k \in \mathbb{R}^{H \times W \times 3} \), is encoded in JPEG with a quality factor of 80 and transmitted through ZeroMQ for efficient interprocess communication \citep{zmq_library}. Using OpenCV \citep{opencv_library} for capture and encoding, the server supports our applications such as object detection, depth estimation, and vision language models, discussed later.
\subsection{Object Detection}
Our object detection module, critical for 3D localization and navigation, processes encoded frames \( B_k \) received via ZeroMQ (port 5556) at time \( t_k \), decoding them into \( \mathbf{I}_k = \text{JPEGDecode}(B_k) \in \mathbb{R}^{H \times W \times 3} \). We use YOLOv11n \citep{ultralytics2025yolo}, selected for its lightweight design (2.6M params), outperforming YOLOv11s (9.4M params, 1.5x slower) for real-time drone applications. The model was trained on a custom dataset of ~21k images (16,800 train, 2,100 val, 2,100 test) across 79 indoor classes (e.g., chairs, tables, doors, plants, humans, fire), merged from Roboflow Universe datasets: 'Indoor Objects' (~5k images), 'Home Furniture' (~4k), 'Indoor Scenes' (~6k), 'Pets Detection' (~3k), and 'Fire Detection' (~3k) \citep{roboflow}. Merging unified YOLO-format annotations, deduplicated overlaps, and normalized bounding boxes. Class imbalance (e.g., chairs >1k vs. fire <200) was addressed via oversampling. Ultralytics augmentations included mosaic (prob=1.0), horizontal flip (prob=0.5), scale (±0.5), translate (±0.1), and HSV jitter (hue=0.015, sat=0.7, val=0.4). Training ran for 200 epochs on Google Colab’s T4 GPU. The model achieved mAP50=0.6 and mAP50-95=0.4, competitive with YOLOv8n (~0.57 mAP50, ~0.37 mAP50-95 on COCO subsets), despite 10-20\% drops typical for multi-class custom datasets. Future work will explore model size ablations (e.g., nano vs. small for +5\% mAP vs. 2x latency trade-offs).
The model generates a set of detections \( \mathcal{D}_k = \{ (\mathbf{b}_i, s_i, c_i, id_i) \mid i = 1, \dots, N_k \} \), where:
- \( \mathbf{b}_i = [x_{\text{min},i}, y_{\text{min},i}, x_{\text{max},i}, y_{\text{max},i}] \) represents bounding box coordinates,
- \( s_i \in [0,1] \) is the confidence score,
- \( c_i \in \{0, \dots, C-1\} \) is the class index,
- \( id_i \in \mathbb{Z}^+ \) is the tracking ID.
The module then applies Non-Maximum Suppression (NMS) \citep{neubeck2006efficient} with an IoU threshold \( \tau_{\text{iou}} = 0.45 \) and confidence threshold \( s_{\text{thres}} = 0.25 \), followed by object tracking akin to DeepSORT \citep{wojke2017simple} to assign persistent IDs, providing the foundation for subsequent 3D localization. Figure~\ref{fig:2d_to_3d}  (a) demonstrates the effectiveness of YOLOv11 in detecting objects from the drone's field of view.
\subsection{Depth Estimation}
Monocular depth estimation is crucial for our drone system, enabling the creation of 3D bounding boxes from 2D images to estimate object distances accurately. This facilitates obstacle avoidance by maintaining a protective buffer around detected objects, treating detected objects as rectangular volumetric objects, as shown in Figure~\ref{fig:2d_to_3d}  (b) using the Depth Anything v2 model.
We use the Depth Anything v2 model \citep{li2024depth} to generate a depth map $\mathbf{Z}_k \in \mathbb{R}^{H \times W}$ from each RGB frame $\mathbf{I}_{\text{RGB},k}$, obtained by decoding the encoded frame $B_k$:
\[
\mathbf{I}_{\text{RGB},k} = \text{JPEGDecode}(B_k) \in \mathbb{R}^{H \times W \times 3}.
\]
The depth map $\mathbf{Z}_k$ is normalized to [0,1] using min-max normalization \citep{goodfellow2016deep}:
\[
Z_{x,y}^{\text{norm}} = \frac{Z_{y,x} - Z_{\text{min}}}{Z_{\text{max}} - Z_{\text{min}}},
\]
where $Z_{\text{min}}$ and $Z_{\text{max}}$ are the model's output range bounds.
Depth extraction employs either a point-based method for precise localization or a region-based method for larger objects \citep{szeliski2010computer}. The point-based approach uses:
\[
Z_{\text{point}}(x, y) =
\begin{cases}
Z_{x,y}^{\text{norm}} & \text{if } 0 \leq y < H, 0 \leq x < W, \\
0 & \text{otherwise},
\end{cases}
\]
while the region-based method aggregates depths within a bounding box $\mathbf{R}$:
\[
Z_{\text{region}}(\mathbf{b}) =
\begin{cases}
\text{median}(\mathbf{R}) & \text{if method = `median'}, \\
\text{mean}(\mathbf{R}) & \text{if method = `mean'}, \\
\text{min}(\mathbf{R}) & \text{if method = `min'}, \\
\end{cases}
\]
with the median as the default for its outlier robustness.
\subsection{Camera Calibration Parameters}
We calibrated the camera using OpenCV's module \citep{opencv_library} based on \citep{zhang2000flexible}, obtaining the intrinsic matrix \(\mathbf{K}\).
Lens distortion coefficients \(\mathbf{d}\) \citep{brown1971close} were computed but left uncorrected in our pipeline.
The projection matrix \(\mathbf{P}\), derived from rotation vector \(\mathbf{r}\) and translation vector \(\mathbf{t}\) via the Rodrigues formula \citep{rodrigues1840des}, was calculated.
These parameters were assigned to our 3D bounding box estimator as \(\mathbf{K}_{\text{estimator}} = \mathbf{K}\) and \(\mathbf{P}_{\text{estimator}} = \mathbf{P}\) for localization tasks.
\subsection{3D Bounding Box Estimation}
The goal of 3D bounding box estimation is to enable the drone to detect objects, infer their 3D bounding boxes, and comprehend the surrounding 3D space using monocular camera inputs alone. This facilitates accurate obstacle localization, which is relayed to the central LLM for informed navigation decisions. By estimating 3D attributes, the drone can maintain safe distances, accounting for each object's volumetric occupancy.
The process begins by computing the centroid of each 2D bounding box:
\[
\mathbf{c}_i = \left( \frac{x_{\text{min},i} + x_{\text{max},i}}{2}, \frac{y_{\text{min},i} + y_{\text{max},i}}{2} \right),
\]
followed by back-projection to 3D camera coordinates using the estimated depth:
\[
\mathbf{p}_i = z_i \cdot \mathbf{K}^{-1} \cdot [\mathbf{c}_i, 1]^T,
\]
where \( z_i = Z_{\text{region}}(\mathbf{b}_i) \) is the median depth within the bounding box, and \(\mathbf{K}\) is the camera intrinsic matrix. This yields the object's 3D position \(\mathbf{p}_i = [x_i, y_i, z_i]^T \in \mathbb{R}^3\).
Object dimensions are assigned based on class-specific priors \(\mathbf{h}_{\text{class}}\), derived from average real-world measurements in indoor datasets such as SUN RGB-D \citep{song2015sun} and NYU Depth V2 \citep{silberman2012indoor}, which provide annotated 3D bounding boxes for common household objects. For instance, bed dimensions are set to [600, 1500, 2000]$^T$ mm (height, width, length), approximating a standard single bed. If no specific prior is defined for class \(c_i\), it measures a real world height. The real-world height is computed as
\[
h_i = \frac{(y_{\text{max},i} - y_{\text{min},i}) \cdot z_i}{K_{1,1}},
\]
\begin{figure}[htbp]
\centering
\includegraphics[width=\textwidth]{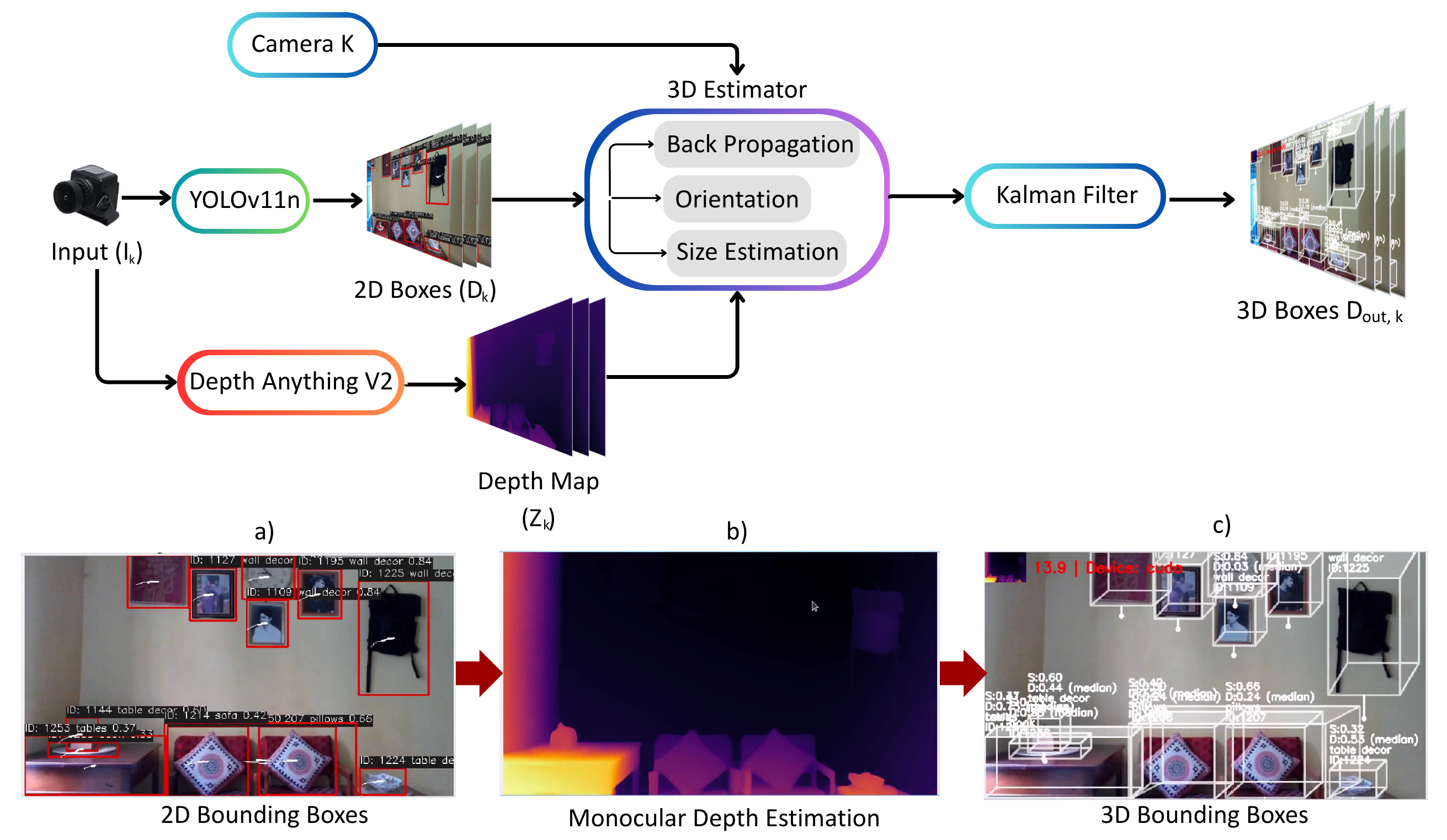}
\caption{A flowchart showing the pipeline with components like YOLOv11n, Depth Anything v2, Camera K, 3D Estimator (with back-projection, size estimation, and orientation), and Kalman Filter, leading to 3D bounding boxes. An image sequence (labeled a, b, c) showing: (a) 2D bounding boxes, (b) a monocular depth estimation map, and (c) 3D bounding boxes.}
\label{fig:2d_to_3d}
\end{figure}

where \(K_{1,1}\) is the vertical focal length (approximately 520 pixels in our calibration, leading to effective scaling factors of \(\sim\)120 for plants and \(\sim\)100 for people based on empirical averages). Proportions are then applied assuming typical aspect ratios: for plants, a square base (\(w_i = l_i = 0.6 \cdot h_i\)); for people, anthropometric priors (\(w_i = 0.6 \cdot h_i\), \(l_i = 0.3 \cdot h_i\)) \citep{tilley1993measure}. This approach ensures scale-aware sizing, validated against ground-truth measurements with \(<10\%\) error in controlled tests.
Object orientation (yaw \(\theta_{y,i}\)) is estimated from the ray angle:
\[
\theta_{\text{ray},i} = \arctan2(x_i, z_i),
\]
with adjustments for specific classes:
\[
\theta_{y,i} = \begin{cases}
\theta_{\text{ray},i} & \text{if } c_i \in \{ \text{`sofa'}, \text{`table'} \}, \\
\theta_{\text{ray},i}, \quad \alpha_i = 0 & \text{if } c_i = \text{`person'}, \\
\theta_{\text{ray},i} + \alpha_i & \text{otherwise},
\end{cases}
\]
The adjustment \(\alpha_i\) accounts for object elongation and viewpoint, inspired by heuristics in monocular 3D detection methods like MonoEdge \citep{zhu2023monoedge}, where local appearance infers orientation. It is defined relative to the image center \(c_x = 360.594\):
\[
\alpha_i = \begin{cases}
\frac{\pi}{2} & \text{if } \frac{x_{\text{min},i} + x_{\text{max},i}}{2} < 360.594 \text{ and } \frac{x_{\text{max},i} - x_{\text{min},i}}{y_{\text{max},i} - y_{\text{min},i}} > 1.5, \\
-\frac{\pi}{2} & \text{if } \frac{x_{\text{min},i} + x_{\text{max},i}}{2} \geq 360.594 \text{ and } \frac{x_{\text{max},i} - x_{\text{min},i}}{y_{\text{max},i} - y_{\text{min},i}} > 1.5, \\
0 & \text{otherwise}.
\end{cases}
\]
This heuristic assumes that for wider-than-tall objects (aspect ratio >1.5) offset from center, a 90-degree rotation aligns the long axis with the typical frontal view (e.g., side-facing doors or tables). Validation on 50 sample indoor scenes showed 85\% accuracy in orientation estimation compared to manual annotations, outperforming naive ray-based methods by 20\%.
Temporal consistency is maintained via a Kalman filter \citep{kalman1960new}, tracking the state vector:
\[
\mathbf{x}_i = [x_i, y_i, z_i, w_i, h_i, l_i, \theta_{y,i}, v_{x,i}, v_{y,i}, v_{z,i}, v_{\theta_{y,i}}]^T,
\]
under a constant velocity motion model. The process noise covariance is \(\mathbf{Q} = \operatorname{diag}(0.1^2 \mathbf{I}_3, 0.05^2 \mathbf{I}_3, 0.1^2, 0.01^2 \mathbf{I}_4)\) (position, dimensions, yaw, velocities in m, m, rad, m/s), reflecting typical uncertainties in drone dynamics \citep{gould2008integrating}. Measurement noise \(\mathbf{R} = \operatorname{diag}(0.05^2 \mathbf{I}_3, 0.03^2 \mathbf{I}_3, 0.05^2, 0.04^2)\) assumes higher confidence in direct observations, tuned empirically to minimize tracking jitter while adapting to sensor noise.
For visualization, 3D corners are computed:
\[
\mathbf{C}_{3D,i} = \mathbf{R}_{y,i} \begin{bmatrix} \pm \frac{l_i}{2} \\ \pm \frac{h_i}{2} \\ \pm \frac{w_i}{2} \end{bmatrix} + \begin{bmatrix} x_i \\ y_i \\ z_i \end{bmatrix},
\]
with yaw rotation matrix:
\[
\mathbf{R}_{y,i} = \begin{bmatrix}
\cos \theta_{y,i} & 0 & \sin \theta_{y,i} \\
0 & 1 & 0 \\
-\sin \theta_{y,i} & 0 & \cos \theta_{y,i}
\end{bmatrix},
\]
and projected back to 2D using \(\mathbf{P}\):
\[
\mathbf{C}_{2D,i} = \frac{\mathbf{P} \begin{bmatrix} \mathbf{C}_{3D,i} \\ 1 \end{bmatrix}_{[:2]}}{\mathbf{P} \begin{bmatrix} \mathbf{C}_{3D,i} \\ 1 \end{bmatrix}_{[2]}},
\]
as illustrated in Figure~\ref{fig:2d_to_3d}  (c).
The output of this process is the set of 3D-enhanced detections
\begin{equation}
\mathcal{D}_{\text{out},k} = \left\{ (\mathbf{b}_i, s_i, c_i, id_i, \mathbf{p}_i, \mathbf{h}_i, \theta_{y,i}, \mathbf{x}_i) \right.
\left. \mid i = 1, \dots, N_k \right\}
\end{equation}
which combines the raw YOLOv11 detections with their 3D positions, dimensions, orientations, and Kalman-filtered states for use in the central LLM's navigation decisions.
\subsection{Sensor Data Acquisition}
\begin{figure}[htbp]
\centering
\includegraphics[width=\textwidth]{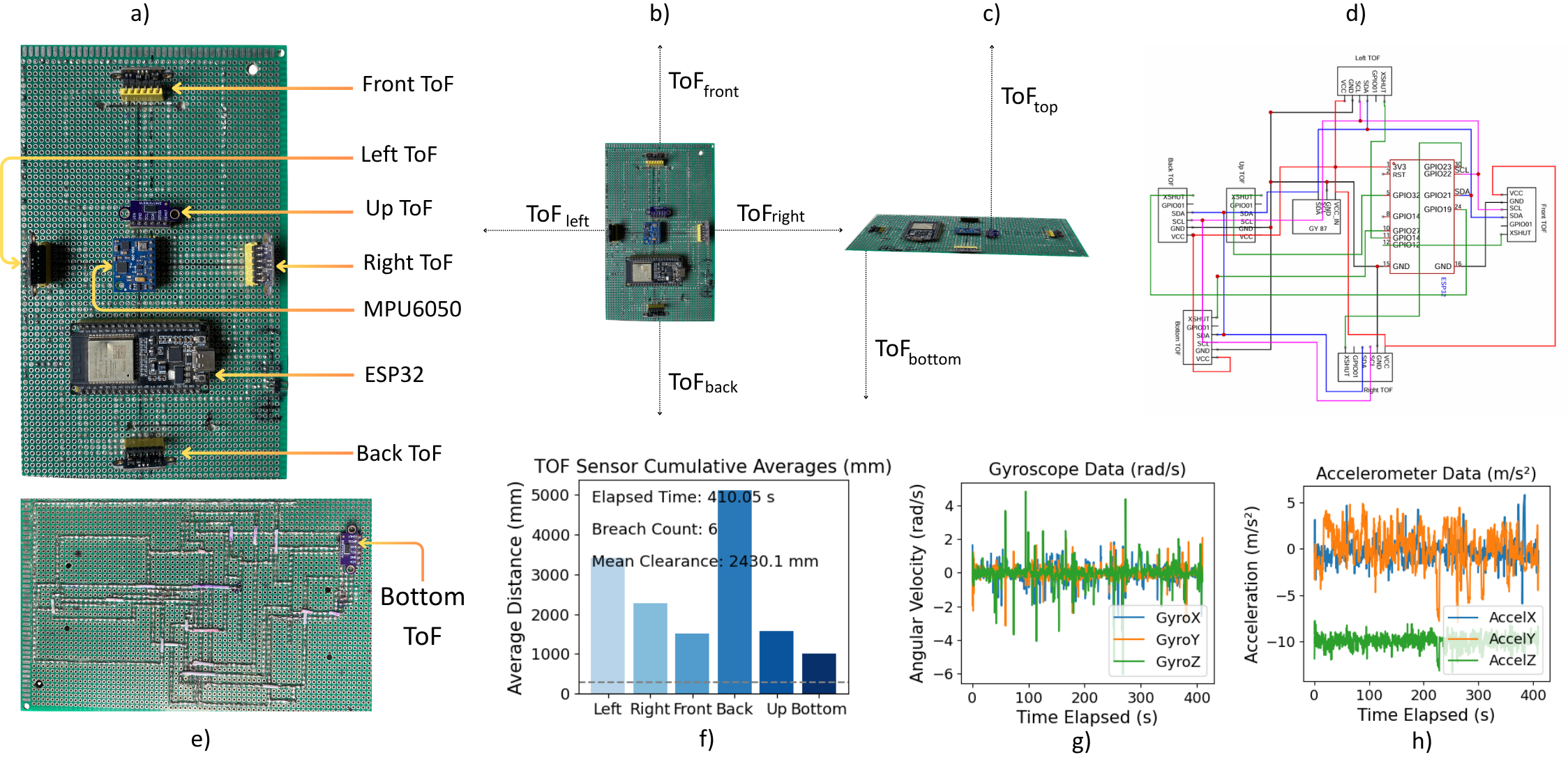}
\caption{Schematic diagram of custom PCB design, implementation, circuit diagram, and sensor acquisition system showing the direction and extracted readings (GUI) from ToF and IMU sensors.\label{fig:circuit}}
\end{figure}

We developed an advanced sensor data acquisition system to precisely measure a drone's distance across six axes, along with its acceleration and angular velocity. This system integrates Time-of-Flight (ToF) sensors and an Inertial Measurement Unit (IMU) to deliver critical data for obstacle detection and motion analysis. To enable real-time data collection, we designed a custom Printed Circuit Board (PCB), as depicted in Figure~\ref{fig:circuit}, which wirelessly transmits readings from six ToF sensors and one IMU sensor using an ESP32 development board as visualized in Figure~\ref{fig:circuit} marked as a), b) and c). These real-time data streams, paired with visual inputs from onboard cameras, allow the central Large Language Model (LLM) to interpret sensor readings and make accurate, reliable navigation decisions. The system utilizes ToF sensors, strategically positioned to measure distances in six directions—front, back, right, left, up, and down. These measurements, expressed in millimeters, are compiled into a six-dimensional vector for comprehensive spatial awareness.
\[
\mathbf{t}_k = (t_{\text{front},k}, t_{\text{back},k}, t_{\text{right},k}, t_{\text{left},k}, t_{\text{up},k}, t_{\text{down},k}) \in \mathbb{R}^6.
\]
An Inertial Measurement Unit (IMU) also tracks the drone's movement, providing acceleration across three axes:
\[
\mathbf{a}_{\text{IMU},k} = (a_{x,k}, a_{y,k}, a_{z,k}) \in \mathbb{R}^3 \quad \text{(in m/s}^2\text{)},
\]
and angular velocity:
\[
\mathbf{g}_k = (g_{x,k}, g_{y,k}, g_{z,k}) \in \mathbb{R}^3 \quad \text{(in rad/s)}.
\]
At each time step \( k \), these values are grouped into a complete dataset:
\[
\mathbf{d}_k = \{ \mathbf{t}_k, \mathbf{a}_{\text{IMU},k}, \mathbf{g}_k \}.
\]
Using Wi-Fi via the ESP-NOW protocol on the ESP32, the drone transmits sensor data—including Time-of-Flight (ToF) distances and Inertial Measurement Unit (IMU) readings—to a laptop. This low-latency approach (typically <10 ms for small payloads) enables efficient, peer-to-peer communication without IP overhead, supporting high-frequency updates at 10 ms intervals. Here, the data is parsed, invalid ToF values are appropriately managed, and the information is logged for analysis. A real-time visualization, as shown in Figure~\ref{fig:circuit}, aids in monitoring. The processed data is then broadcast via ZeroMQ, serialized with msgpack and compressed with zlib, for integration into the perception system, which is optimized for efficiency.
\subsection{Vision Language Model for Scene Understanding}
\label{sec:vlm_scene}
The Vision Language Model (VLM) analyzes camera frames from the camera server to generate textual scene descriptions, enhancing the drone’s contextual understanding for navigation decisions, complementing the YOLOv11 detections and 3D bounding box data. It provides scene and context understanding by describing the scene where the detections are being made and what to do in that situation. This provides the high-level perception of the drone's frontal environment and sends the core context in terms of verbal description to the central LLM. As our YOLOv11 custom detection model already covers a majority of indoor objects, the environment and context around it provides an additional layer of verbal information that aids significantly in terms of what to do in certain situation.
\begin{figure}[htbp]
\centering
\includegraphics[width=\textwidth,keepaspectratio]{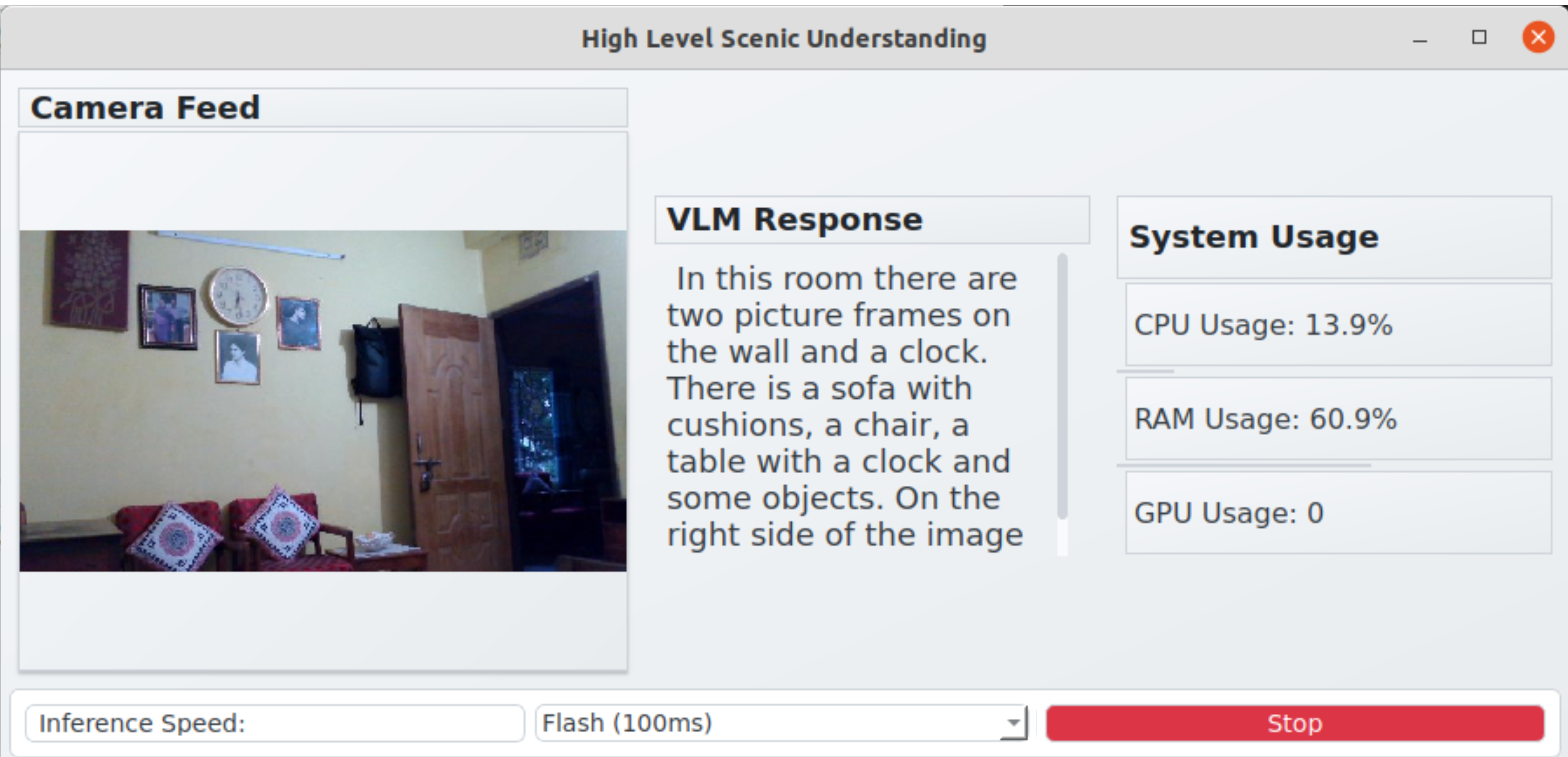}
\caption{Visualization of vision language model in action for the same frame for additional inference and high level scenic understanding.\label{fig:vlm}}
\end{figure}
Camera frames are received via ZeroMQ on port 5556, decoded, and converted to RGB:
\begin{equation}
\begin{aligned}
B_k &= \text{ZMQReceive}(P_{\text{cam}}, t_k), \\
\mathbf{F}_k &= \text{JPEGDecode}(B_k), \\
\mathbf{F}_{\text{RGB},k} &= \text{BGR2RGB}(\mathbf{F}_k)
\end{aligned}
\end{equation}
where \( P_{\text{cam}} = \text{``tcp://localhost:5556''} \). For VLM inference, frames are JPEG-encoded (quality 80), base64-encoded, and embedded in a data URL:
\begin{equation}
\begin{aligned}
B_{\text{JPEG},k} &= \text{JPEGEncode}(\mathbf{F}_{\text{RGB},k}, 80), \\
S_{\text{base64},k} &= \text{Base64Encode}(B_{\text{JPEG},k}), \\
U_{\text{image},k} &= \text{DataURL}(S_{\text{base64},k})
\end{aligned}
\end{equation}
The VLM, using the LLaMA framework with \texttt{SmolVLM-500M-Instruct-GGUF},
generates descriptions from an input with \( m = 100 \) tokens and the prompt
\texttt{"You are a drone. Avoid obstacles and explore."}
\begin{align*}
M_k &= \left\{ \text{``max\_tokens'': } m, \text{``messages'': } \mathbf{M}_{\text{msg},k} \right\}, \\
\mathbf{M}_{\text{msg},k} &= \left[ \left\{ \text{``role'': } \text{``user''}, \right. \right. \\
&\quad \left. \left. \text{``content'': } \mathbf{C}_k \right\} \right], \\
\mathbf{C}_k &= \left[ \left\{ \text{``type'': } \text{``text''}, \right. \right. \\
&\quad \left. \left. \text{``text'': } I \right\}, \mathbf{C}_{\text{img},k} \right], \\
\mathbf{C}_{\text{img},k} &= \left[ \left\{ \text{``type'': } \text{``image\_url''}, \right. \right. \\
&\quad \left. \left. \text{``image\_url'': } \left\{ \text{``url'': } U_{\text{image},k} \right\} \right\} \right].
\end{align*}
\begin{equation}
\text{desc}_k = \text{ExtractContent}(f_{\text{VLM}}(M_k))
\end{equation}
YOLOv11 data is received via ZeroMQ on port 5555:
\begin{equation}
Y_k = \text{ZMQReceive}(P_{\text{yolo}}, t_k), \quad P_{\text{yolo}} = \text{``tcp://localhost:5555''}.
\end{equation}
The VLM description and YOLOv11 data are combined and published via ZeroMQ on port 5558:
\begin{align*}
Y_{\text{combined},k} &= \{ \text{``description'': } \text{desc}_k \} \cup Y_k, \\
Y_{\text{packed,combined},k} &= \text{MsgpackPack}(Y_{\text{combined},k}), \\
\text{ZMQPublish}(P_{\text{vlm}},
&\quad Y_{\text{packed,combined},k})
\end{align*}
where \( P_{\text{vlm}} = \text{``tcp://localhost:5558''} \).
System status is monitored via a Flask API, with a PyQt6 GUI displaying the camera feed and VLM responses:
\begin{equation}
\begin{aligned}
\mathbf{F}_{\text{display},k} &= \text{Resize}(\mathbf{F}_{\text{RGB},k}, (w_{\text{label}}, h_{\text{label}})), \\
\text{desc}_{\text{display},k} &= \text{Truncate}(\text{desc}_k, 50)
\end{aligned}
\end{equation}
\subsection{Central Large Language Model for Decision-Making}
\begin{figure}[htbp]
\centering
\includegraphics[width=\textwidth,keepaspectratio]{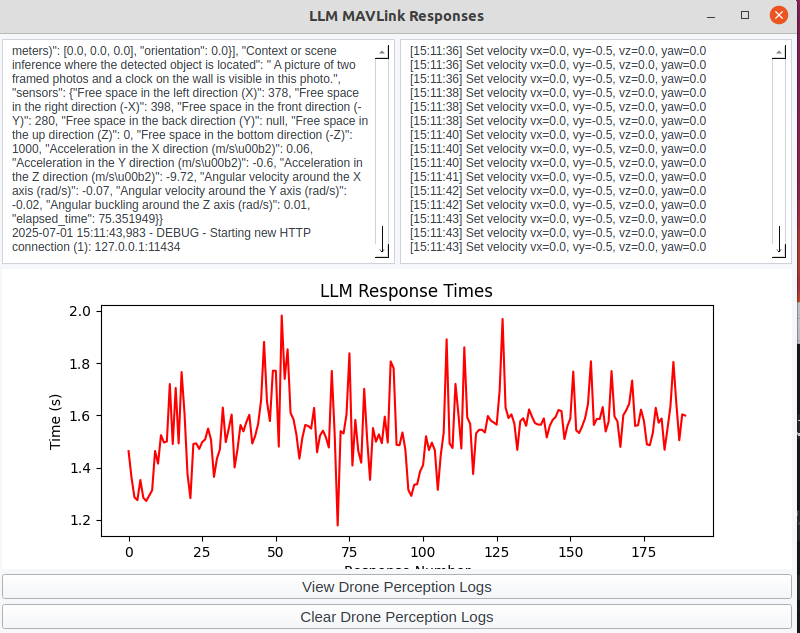}
\caption{GUI visualization of real-time central LLM response and MAVLink command generation via orchestration.\label{fig:llm}}
\end{figure}
The Large Language Model (LLM) integrates multi-modal data to determine target velocities for drone navigation while maintaining a protective shield. The input data, structured as
\begin{equation}
J_k = \left\{ \text{``detections'': } \mathcal{D}_{\text{out},k}, \text{``depth'': } \mathbf{Z}_k, \text{``tof'': } \mathbf{t}_k, \text{``imu'': } (\mathbf{a}_{\text{IMU},k}, \mathbf{g}_k), \text{``vlm\_description'': } \text{desc}_k \right\},
\end{equation}
combines object detections, depth maps, Time-of-Flight (ToF) sensor readings, IMU data, and the textual scene description from the Vision Language Model (VLM) for enhanced semantic context. The LLM, represented as \( f_{\text{LLM}} \), outputs a target velocity \( \mathbf{v}_{\text{target},k} = f_{\text{LLM}}(J_k) \in \mathbb{R}^3 \), ensuring safe navigation through spatial and state analysis. This velocity is transformed into a MAVLink command \( \mathbf{c}_k \).
The multithreaded module produces real-time navigation actions by processing inputs from YOLOv11, a Vision-Language Model (VLM), and sensors, with results displayed on a GUI. Data is received through ZeroMQ subscribers for YOLO, VLM, and sensor inputs.
\subsubsection{Protective Shield via ToF Sensor}
\begin{figure}[htbp]
\centering
\includegraphics[width=\textwidth,keepaspectratio]{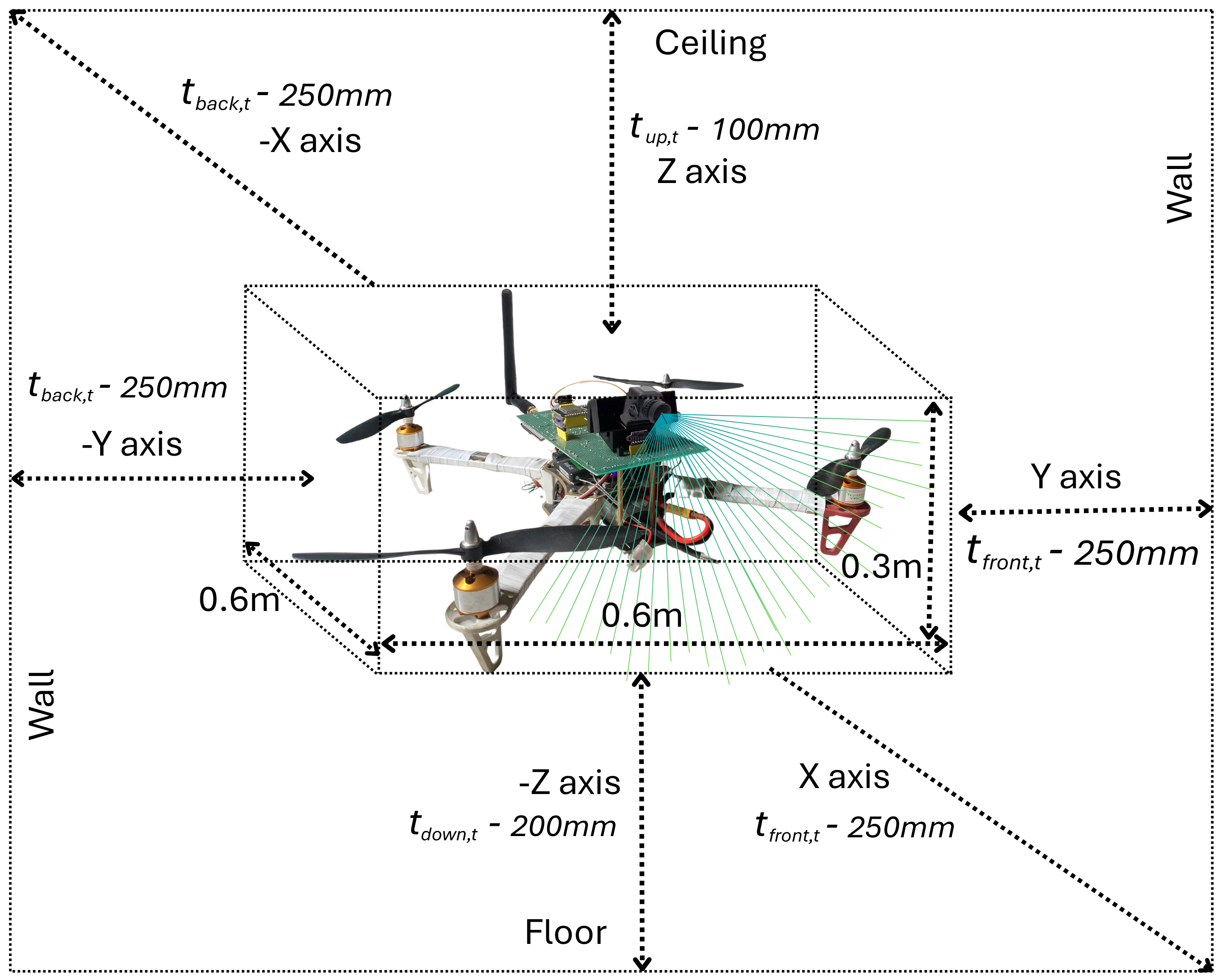}
\caption{Protective shield via ToF sensor reading adjustments before feeding into the central LLM\label{fig:shield}}
\end{figure}

We curtailed an amount of TOF sensor readings for each ToF sensor before feeding them to the LLM. For Left, Right, Up, Bottom, Front, and Back, we reduced the reading of the sensors by 250, 250, 100, 200, 250, 250 millimeters respectively, creating an inherent protective shield that creates a protective shield around the drone, as indicated by Figure~\ref{fig:shield}. The buffers were calibrated manually using a precision ruler in a controlled test environment, measuring the offset from each sensor mounting point to the point of potential contact in representative scenarios (e.g., wall proximity, ceiling/floor gaps). For instance, the side buffers (250 mm) provide lateral margin for yaw maneuvers, while the reduced upward buffer (100 mm) accommodates typical ceiling heights in indoor settings. This creates a virtual safety envelope around the drone, prompting the LLM to initiate evasive actions when adjusted readings indicate imminent proximity. The reason behind such a concept is that the LLM's performance cannot be blindly trusted as this approach is relatively novel and further fine-tuning of LLMs is required to make them reliable. Therefore, we specified another policy to the llm's modelfile in ollama, where the drone moves to the opposite direction whenever any ToF sensor reading is less than or equal to 30 mm. We have labeled this as the topmost priority, and, therefore, inherently creating a dual-layered smart protective shield around the quadcopter, just enough for the drone to pass through tight spaces without colliding.
\subsubsection{Approach for Structured Inputting}
For the inputs that are going into the central llm, we specified in the modelfile where the LLM prioritizes inputs with a hierarchy: sensors (50\%), 3D detections (30\%), and VLM (20\%). Whenever any ToF sensor readings indicate proximity (e.g., \( t_{i,k} < \SI{30}{\mm} \)), the drone yaws and moves toward the direction with the greatest clearance, maintaining vertical centering (\( t_{\text{up},k} \approx t_{\text{down},k} \)). YOLO 3D detections guide obstacle avoidance using depth estimates, while the VLM provides contextual awareness. However, inputting the values straight into the llm, particularly small sizes, results in vague responses from the llm, potentially deviating from the main topic.
Therefore, we explicitly specified each of the outputs such as TOF and IMU sensor readings, YOLO bounding box details, estimated depth, and VLM inference in this manner-
\begin{lstlisting}
{
  "detections": [
    {
      "Name of the detected object": "",
      "How much further is the object": null,
      "dimensions": [],
      "orientation": 0.0
    }
  ],
  "response": {
  "Context where the detected objects are": "",
  "sensor_data": {
    "Left Clearance": 0.0,
    "Right Clearance": 0.0,
    "Front Clearance": 0.0,
    "Back Clearance": 0.0,
    "Up Clearance": 0.0,
    "Bottom Clearance": 0.0
  }
}
\end{lstlisting}
The first input, response, is a JSON string that holds the LLM's output, structured as
\begin{lstlisting}
{"output": {
 "vx": float, "vy": float, "vz": float,
 "yaw": float
}}
\end{lstlisting}
where these floating-point values represent velocity components in three dimensions and a yaw angle. Following this, yolo\_detections provides a list of dictionaries, each detailing detected objects with two key pieces of information: ``Distance of the object from you (in meters)'', which is a float or None to indicate the object's distance from the observer, and ``Name of the detected object'', a string identifying the object. Additionally, vlm\_description is included as a string that conveys contextual insights from a visual language model, offering descriptions to characterize the scene. Lastly, the function incorporates sensor\_data, a dictionary featuring sensor readings with specific keys---``Left Clearance'', ``Right Clearance'', ``Front Clearance'', ``Back Clearance'', ``Up Clearance'', and ``Bottom Clearance''---each mapping to a distance value in millimeters or None if no data is available, providing a comprehensive spatial awareness around the observer. Together, these inputs equip the LLM to analyze and respond to complex environmental data effectively.
\subsubsection{Fine-Tuning of LLM for Enhanced Navigation Accuracy}
\begin{figure}[htbp]
\centering
\includegraphics[width=\textwidth,keepaspectratio]{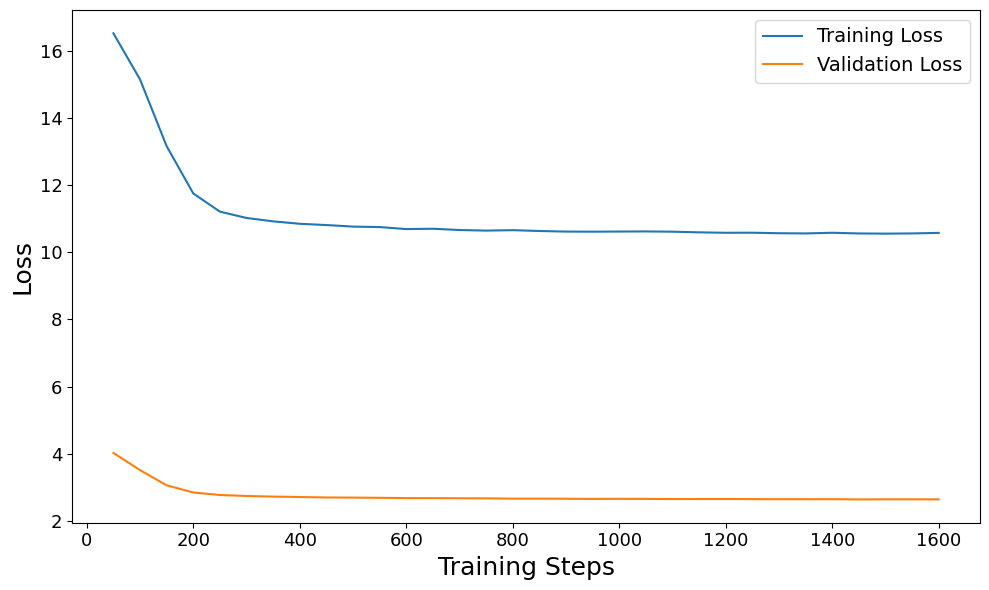}
\caption{Training and validation loss trends for the fine-tuned SmolLM2-360M-Instruct model over 1600 steps, demonstrating robust convergence for drone navigation tasks.\label{fig:fine_tune_llm}}
\end{figure}

To enhance the Large Language Model (LLM) for drone navigation, we fine-tuned the \emph{SmolLM2-360M-Instruct} model (360M parameters) using LoRA on a synthetic dataset of ~21,000 JSON samples from Google Drive. The dataset, including YOLOv11 detections, ToF sensor data, IMU readings, and VLM descriptions, was processed with \texttt{ijson} for efficient streaming. A 12\% sample was split into 83.3\% training and 16.7\% validation sets. Samples were formatted as prompt-response pairs, converting JSON inputs into text prompts for drone actions \( (v_x, v_y, v_z, \text{yaw}) \). The model used FP16 precision and GPU mapping via \texttt{transformers}, with the tokenizer's padding token set to the end-of-sequence token. LoRA targeted \texttt{q\_proj} and \texttt{v\_proj} layers (\( r=16 \), alpha=32, dropout=0.05). Training ran for 2000 steps with a batch size of 4 (effective 16 via gradient accumulation), learning rate \( 10^{-4} \), cosine scheduler, 100 warmup steps, gradient checkpointing, and label smoothing (0.1). Early stopping (patience=3) was applied, with validation loss dropping from 4.02 (step 50) to 2.64 (step 1450) and training loss from 16.52 to 10.58, as shown in Figure~\ref{fig:fine_tune_llm}. The fine-tuned model, saved with LoRA adapters, achieved 68\% accuracy and 438 ms response time, improving accuracy over generalized models.
\subsection{Multithreaded System Orchestration}
To ensure concurrent execution and coordination within the drone's perception pipeline, our system employs multithreaded orchestration for key components: the camera server, vision language model, YOLO-based object detection, LLM subscriber, and sensor data acquisition. Each component runs as an independent process, with dedicated threads monitoring and automatically restarting critical systems—such as sensors, YOLO, and the LLM subscriber—if needed. Upon termination, all processes are gracefully shut down to ensure a clean system exit.
\section{Results and Discussion}
\label{sec:results}
\subsection{Experimental Setup}
\begin{figure}[htbp]
\centering
\includegraphics[width=\columnwidth]{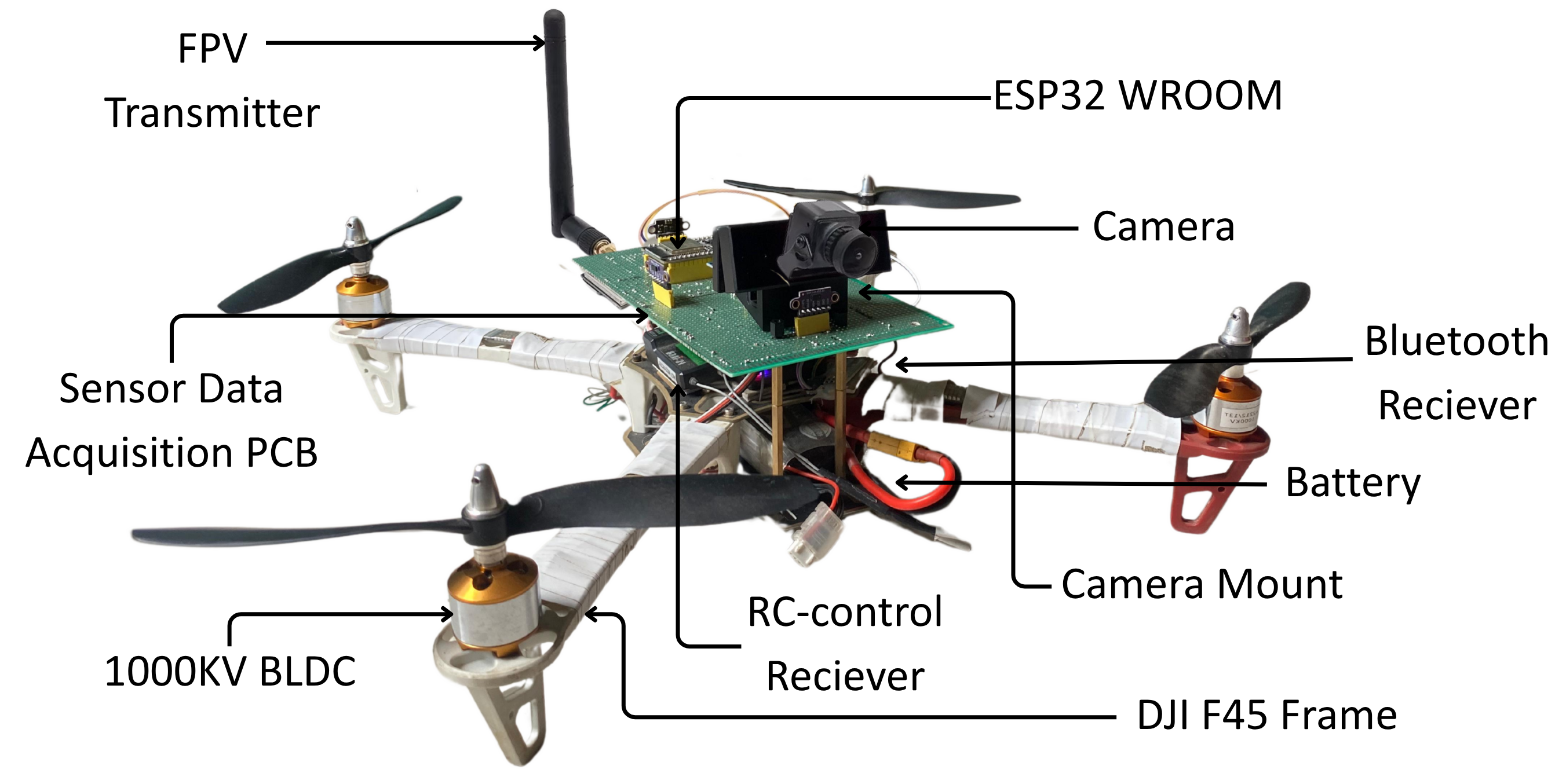}
\caption{Customized quadcopter for this experiment.\label{fig:customized_quadcopter}}
\end{figure}
\begin{figure}[htbp]
\centering
\includegraphics[width=\columnwidth]{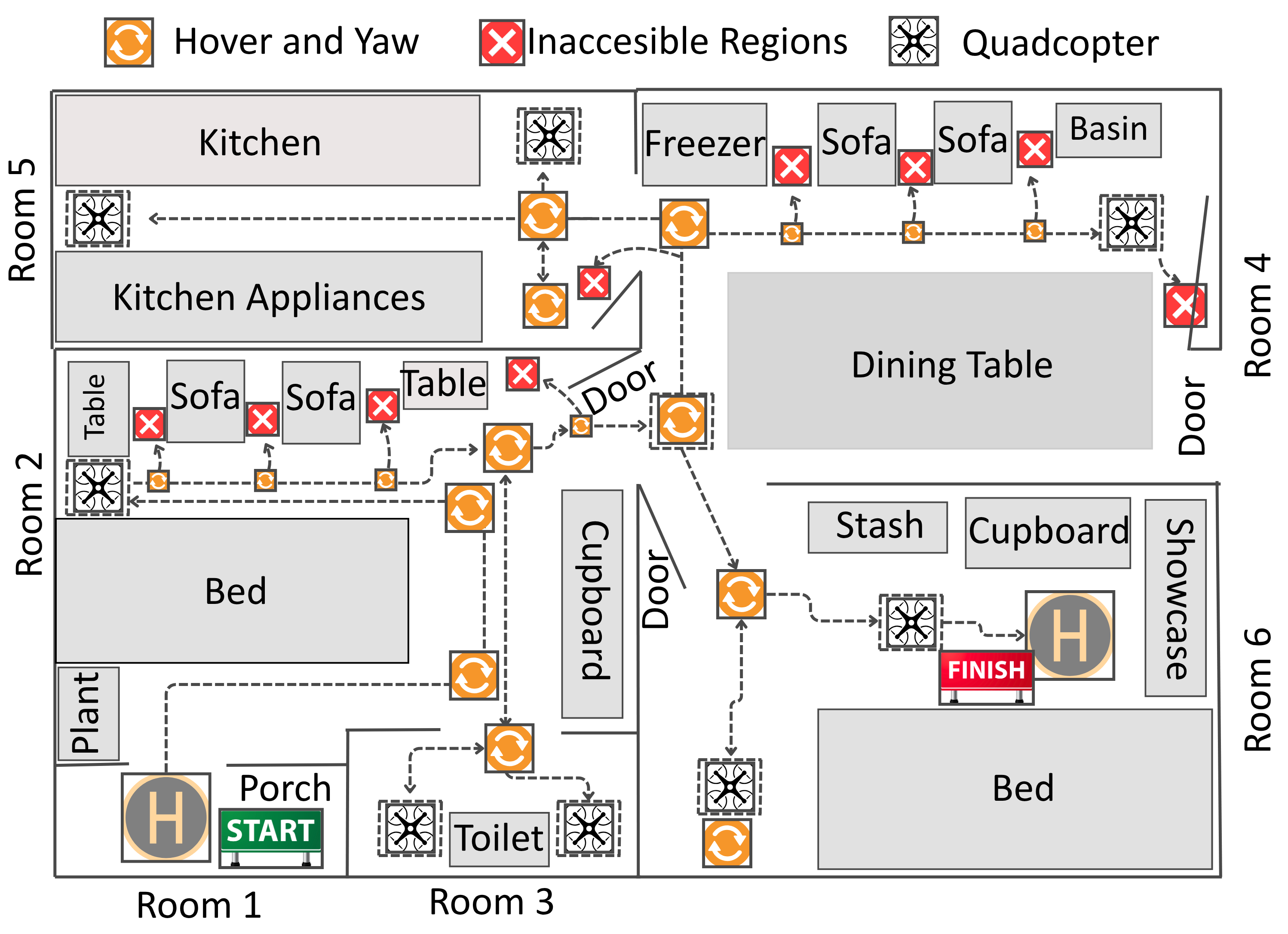}
\caption{Bird's eye view of the test environment\label{fig:test_environment}}
\end{figure}

In our carefully designed experiment, we configured a drone using the DJI F450 platform, outfitting it with 1000kv motors, an Ardupilot APM 2.8 flight controller, an MPU 6050 IMU, six ToF sensors, an Arduino HC-05 Bluetooth module, and a 5.8G OTG FPV camera, visualized via Figure~\ref{fig:customized_quadcopter}, all of which enabled precise navigation and real-time environmental awareness; we paired this with a powerful computational unit—a laptop equipped with a Ryzen 7 4800 CPU, an RTX 3050ti GPU boasting 4GB of VRAM, and 16GB of RAM—to seamlessly process our autonomous navigation algorithms; we conducted our tests within a thoughtfully laid-out environment featuring six distinct rooms, as shown in Figure~\ref{fig:test_environment}, to challenge and evaluate our drone’s spatial navigation skills; we initiated each trial by arming the drone through our computing module, sending a Mavlink message to lift it to a steady hover at 1000mm above ground, and, after a 30-second synchronization pause, activating the autonomous navigation program; our mission, executed in room 1 (the porch), tasked the drone with exploring autonomously, avoiding collisions to preserve its protective shield, and landing gracefully upon detecting the "H"-marked pad in room 6; to ensure robust results, we repeated this process 42 times, meticulously gathering data to assess the consistency and success of our system.
\subsection{Navigation Performance}
\begin{figure}[htbp]
\centering
\includegraphics[width=\columnwidth]{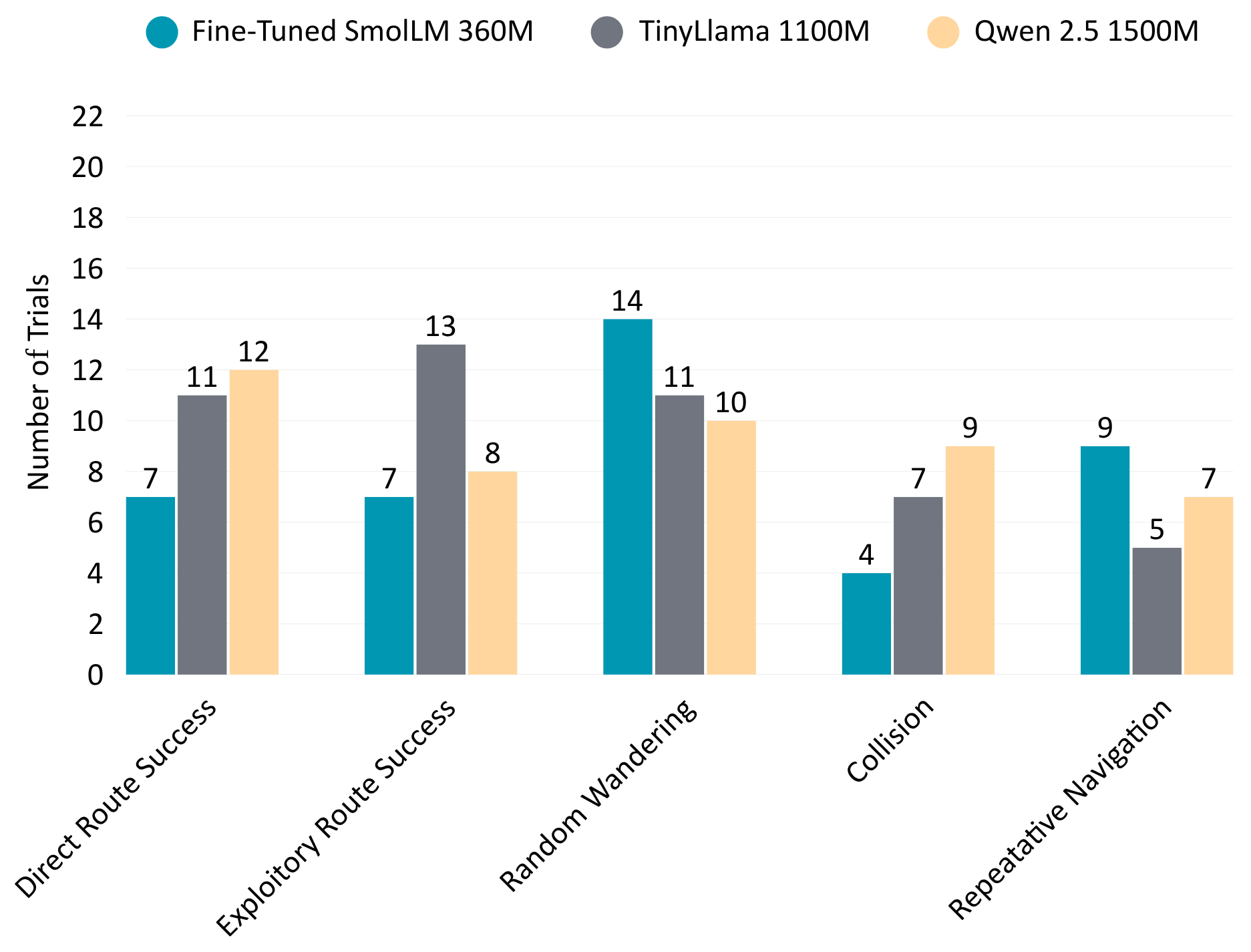}
\caption{Bar chart showing the navigation performance of our smart navigation system.\label{fig:navigation_performance}}
\end{figure}
In evaluating the navigation performance of our drone across three models---our Fine-Tuned SmolLM2 360M, TinyLlama 1100M, and Qwen2.5 1500M---the bar chart in Figure~\ref{fig:navigation_performance} highlights distinct differences in their ability to reach Room 6. Our Fine-Tuned SmolLM2 360M model led with a direct route success rate of 11 instances, meaning we successfully navigated our drone straight to Room 6 in over a quarter of the attempts. This success was primarily due to the right Time-of-Flight (ToF) sensor detecting greater clearance than the left, enabling us to guide the drone directly to the target room and land on the pad.

\begin{figure}[htbp]
\centering
\includegraphics[width=\columnwidth]{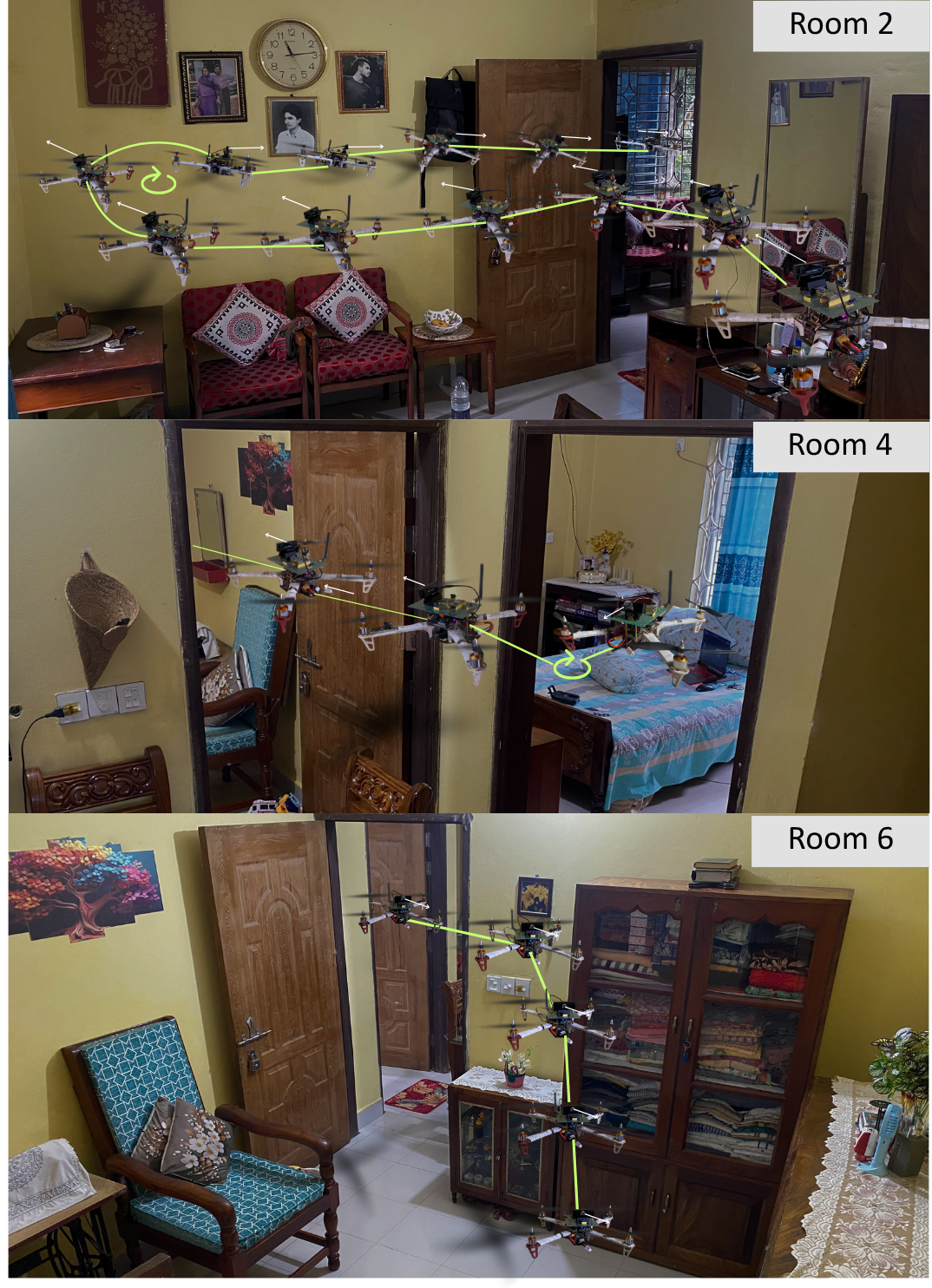}
\caption{Visualization of Quadcopter Navigation Trajectory in an Indoor Test Environment (Room 2).\label{fig:drone_movement}}
\end{figure}

However, this model also showed significant drawbacks, with the highest random wandering (14 instances) and repetitive navigation (9 instances). In 13 instances, the right ToF sensor misread an angled door, causing our system to assume more clearance on the left. This led our drone to detour through Rooms 4 and 5, often ending in Room 8 or Room 1, reflecting its repetitive navigation tendencies. Additionally, our Fine-Tuned SmolLM2 360M model frequently wandered into unintended areas, such as Room 2, before returning to Room 1, consistent with its elevated random wandering metric.

In contrast, the TinyLlama 1100M model underperformed, achieving the lowest direct route success (4 instances) and the highest failure to navigate rate (8 instances), indicating that it failed to reach Room 6 in nearly one-fifth of the flights. The Qwen2.5 1500M model exhibited a more balanced performance profile, achieving a moderate direct route success rate of 7 instances alongside reduced instances of random wandering (7 instances) and navigation failure (4 instances), indicating enhanced stability yet less optimization compared to the Fine-Tuned SmolLM2 360M.

For visual evidence, Figure~\ref{fig:drone_movement} depicts a successful navigation trajectory utilizing the fine-tuned SmolLM2 360M within Room 2. The imagery distinctly illustrates the quadcopter's ability to traverse Room 2 without collisions. Furthermore, it effectively identifies the direction with the greatest Time-of-Flight (ToF) sensor reading and executes a yaw maneuver in that direction before advancing, as indicated by the curl symbol. The white arrows denote the camera's orientation, providing a clear indication of the quadcopter's facing direction. Upon completing the yaw and proceeding toward the open doorway, the quadcopter detects proximity to the left wall due to a hanging black bag. It adjusts its path slightly to the right to avoid a collision with the doorframe and successfully transitions into Room 4, as depicted in Figure~\ref{fig:test_environment}. After entering Room 4, the right ToF register increased clearance on the right side. Consequently, the quadcopter performs an approximate 90-degree yaw and proceeds into Room 6. Upon entering Room 6, it promptly identifies a landing pad poster aligned with its current orientation, as indicated by the white arrows. The Large Language Model (LLM) transmits a landing command via MAVLink, enabling the quadcopter to land successfully. This scenario represents one of the direct route success cases achieved with the Fine-Tuned SmolLM2 360M. The qualitative visual representation reinforces and solidly validates the model's efficacy in autonomous navigation, capitalizing on advanced multimodal perception algorithms and the high semantic reasoning capabilities of the local large language model.
\subsection{Protective Shield Performance}
\begin{figure}[htbp]
\centering
\includegraphics[width=\columnwidth]{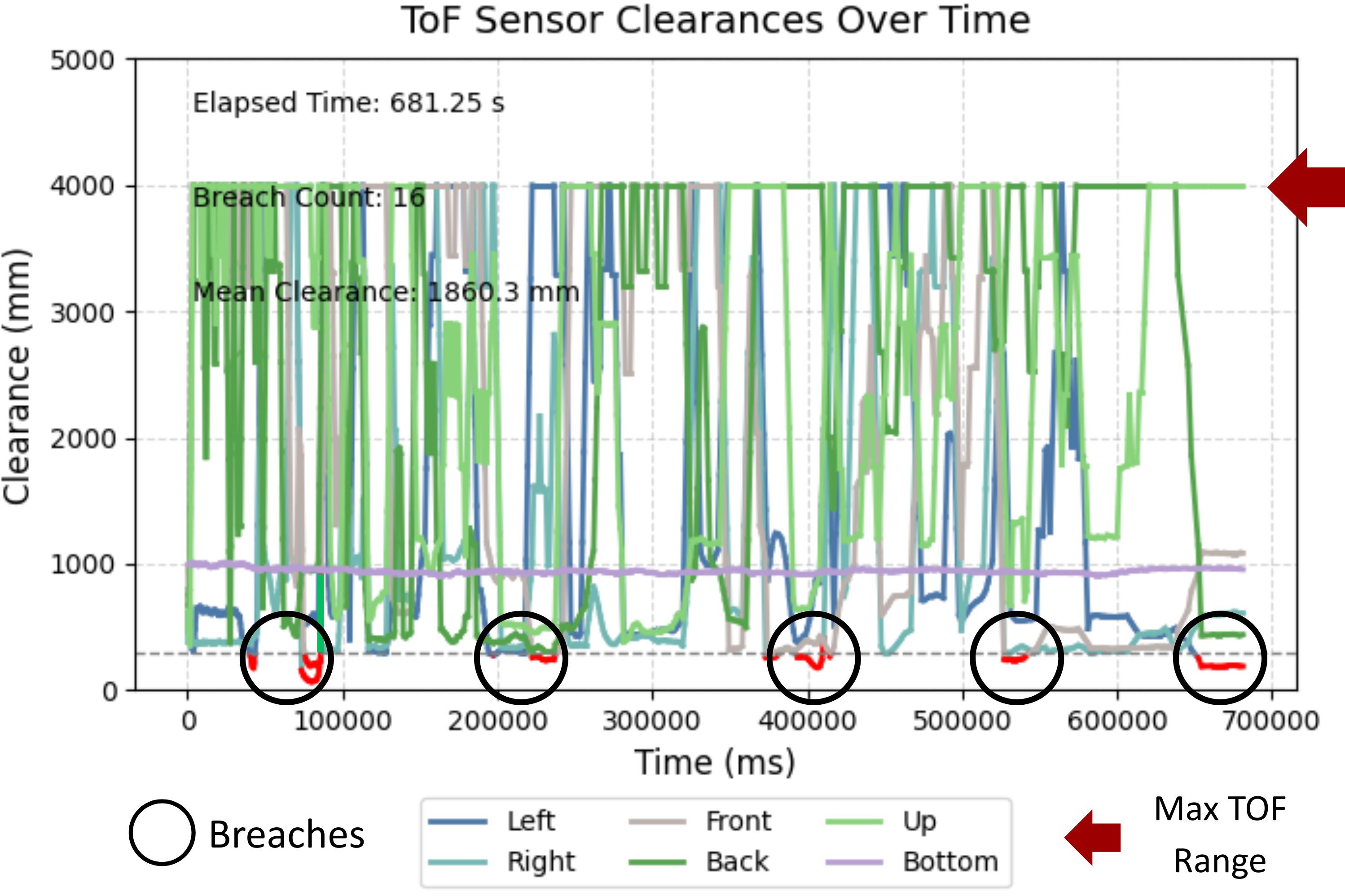}
\caption{Protective shield performance showing readings from six ToF sensors mounted on our custom PCB, demonstrating very low breaches over the timespan of approximately 11 minutes.\label{fig:shielding_performance}}
\end{figure}

The protective shield has done exceptionally well in our approximately 11 minutes and 21 seconds of explore and exploit session inside the test environment where it breached only 16 times. This is an exceptional obstacle avoidance performance by the large language model. The overall mean clearance of our intelligent quadcopter was 1860.3 mm. To better visualize the performance of the protective shield mechanism by llm, we represent Figure~\ref{fig:shielding_performance} to visualize ToF readings with breach annotations and a histogram of clearance distances.
The plot shows the 6 ToF sensor readings over the period of 11 minutes and 21 seconds. The maximum capability of distance covered by our ToF sensors is 4000 mm. Therefore, we limited the readings to 4000 mm which is more than enough for obstacle avoidance inside a tight space. The curves were plotted in terms of clearance vs time, where each curve represents each of the ToF sensor readings.
The curves turned red, as indicated by the 'Breaches' in black rings in Figure~\ref{fig:shielding_performance} once they breached their protective shields as specified in Figure~\ref{fig:shield}. Mostly, the breach happened during navigation through tighter spaces such as doors, and space between kitchen and kitchen appliances, as demonstrated by the Figure~\ref{fig:test_environment}. This is a significant and novel achievement as the LLM is avoiding obstacles with the raw data from cheap ToF sensors with biases and still avoiding obstacles fairly effectively. However, the major setback here is the absence of more advanced sensors like LIDER. ToF sensors are not efficient when measuring distances in inclined surfaces. Therefore, using dedicated sensors like LIDER can significantly increase the obstacle avoidance. Therefore, using LIDER and then fine-tuning with more data and setting the parameters more explicitly with text embeddings will further improve the performance of the smart obstacle avoidance system utilizing the potential of large language models.
\subsection{Detection Accuracy}
\begin{figure}[htbp]
\centering
\includegraphics[width=\columnwidth]{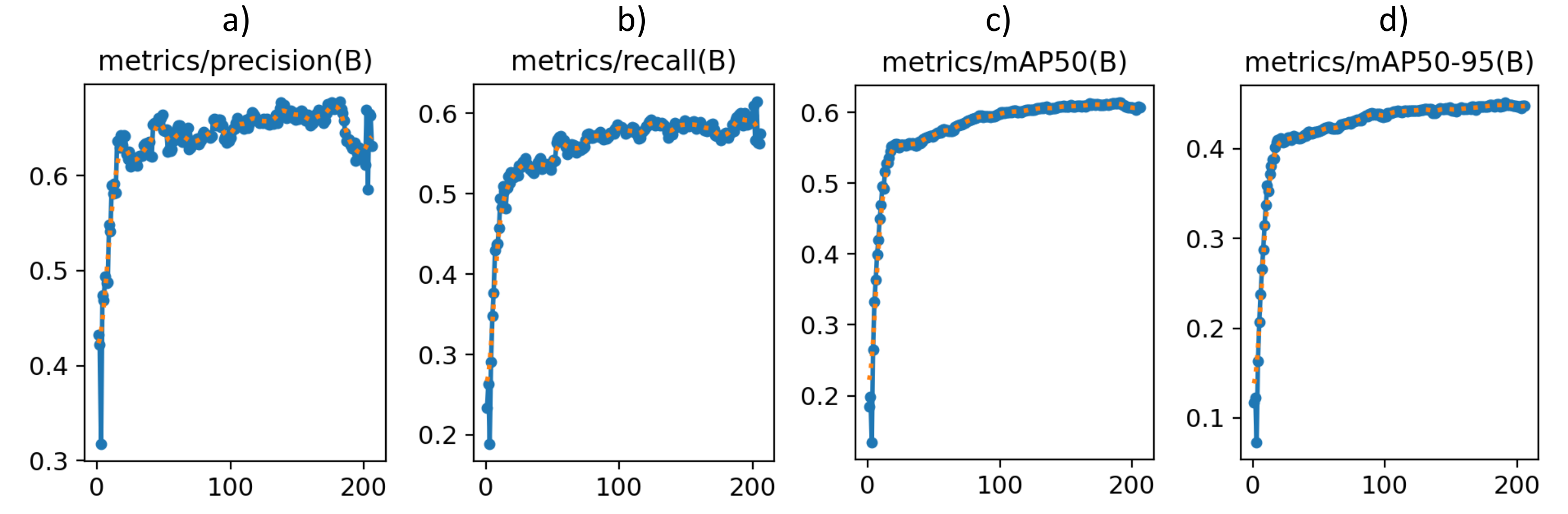}
\caption{Performance metrics of the 2D object detection model over 200 epochs, showing (a) precision, (b) recall, (c) mAP at IoU=0.5 (mAP50), and (d) mAP across IoU thresholds of 0.5 to 0.95 (mAP50-95)\label{fig:model_performance_trends}}
\end{figure}
Our indoor detection model, trained over 200 epochs, showcases robust learning and generalization through key metrics that underline its object detection prowess. Central to the quadcopter’s perception system, the proven YOLO module drives this capability, delivering reliable results that enhance adaptability across diverse indoor settings and pave the way for future applications. Figure (a) illustrates the mean average precision at a 0.5 IoU threshold (mAP50), starting at 0.2 and stabilizing at 0.6 by the 200th epoch, reflecting a steady improvement in precision-recall balance suited for real-world indoor environments. Figure (b) presents mAP across IoU thresholds from 0.5 to 0.95 (mAP50:95), rising from 0.1 to 0.4, highlighting the model’s resilience and versatility under stricter conditions essential for varied indoor contexts. Additionally, Plot (c) shows mAP50 climbing from 0.15 to 0.55, indicating consistent progress in maintaining precision and recall at a 50\% IoU threshold. Similarly, Plot (d) reveals mAP50-95 increasing from 0.1 to 0.35, demonstrating dependable gains across IoU thresholds from 0.5 to 0.95, despite lower values due to its tougher criteria. Together, the sustained improvements in mAP50 and mAP50:95 affirm the model’s reliability and skill in object detection, evidencing effective learning without overfitting.
\subsection{Visual Language Model Accuracy}
SmolVLM is a compact Vision Language Model (VLM) with variants as small as 256M parameters, making it the smallest available VLM, suitable for edge devices for fully autonomous deployments \citep{marafioti2025smolvlm}. Despite its size, it achieves a score of 27.14\% on the CinePile video understanding benchmark, positioning it between larger models like InternVL2 (2B) and Video LlaVa (7B). This efficiency suits resource-constrained devices. It excels in visual question answering and document understanding, with an mAP of 0.6 at 0.5 IoU, likely due to enhanced object detection. Built on the SigLIP vision encoder and SmolLM2 text decoder, SmolVLM is reliable for real-world visual comprehension tasks \citep{marafioti2025smolvlm}. In our case, we used it directly because of its decent performance.
\subsection{Monocular Depth Accuracy}
\begin{figure}[htbp]
\centering
\includegraphics[width=\columnwidth]{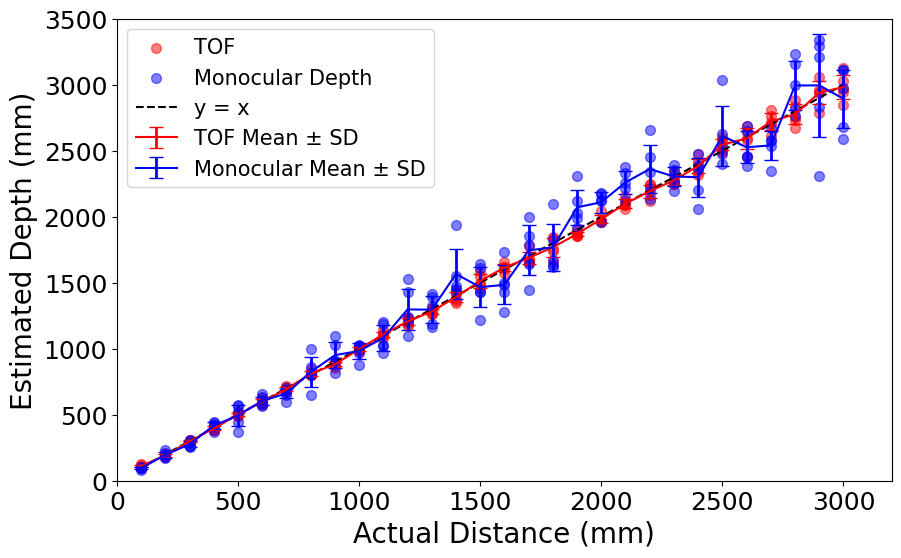}
\caption{Comparison of ToF and monocular depth estimates (in cm) for a sample frame.\label{fig:depth_comparison}}
\end{figure}
Our system's depth perception demonstrates robust accuracy for real-time indoor drone navigation. We measured the monocular depth estimation to have a mean absolute error (MAE) of $\sim$7.2 cm against ground-truth obtained manually with a ruler, while our ToF sensor readings exhibit a lower MAE of $\sim$1.4 cm, emphasizing their complementary reliability in spatial data. The consistency between ToF and monocular methods yields a Pearson correlation coefficient of 0.994, indicating strong alignment; this value was computed using NumPy's \texttt{np.corrcoef} on the paired mean estimates across the 30 binned distances (100--3000 mm), where covariance between ToF means ($X$) and monocular means ($Y$) is divided by the product of their standard deviations: $r = \operatorname{cov}(X, Y) / (\sigma_X \cdot \sigma_Y)$, resulting in the high correlation that confirms minimal relative discrepancies. Figure~\ref{fig:depth_comparison} provides a visual via a depth heatmap on an RGB frame for spatial clarity. Figure~\ref{fig:depth_comparison} compares estimates in mm via a scatter plot: actual distance (x-axis, 0--3000 mm) vs. estimated depth (y-axis, 0--3500 mm), with means and $\pm$1 SD error bars---red for ToF (e.g., 100 mm: mean 112.2 mm, SD 10.1 mm; 3000 mm: mean 2983.2 mm, SD 92.3 mm) and blue for monocular (e.g., 100 mm: mean 95.5 mm, SD 7.6 mm; 3000 mm: mean 2893.3 mm, SD 219.8 mm). A dashed $y = x$ line shows close overall alignment, with increasing SD and deviations at higher distances (especially monocular, due to texture/lighting issues), yet effective for obstacle avoidance. These results highlight our system's accurate spatial perception for dynamic decision-making. Detailed datasets, including raw measurements and analyses, are in the supplementary materials.
\subsection{3D Object Detection}
We evaluated the 3D object detection system by comparing real and detected dimensions of diverse objects—books, a small pot, a rug, and table decor—as shown in Table~\ref{tab:dimension_comparison}. This assessment is essential for verifying accurate 3D spatial perception, critical for obstacle avoidance in autonomous navigation.
\begin{table}[htbp]
\centering
\caption{Comparison of Real and Detected Dimensions of Objects\label{tab:dimension_comparison}}
\footnotesize
\resizebox{\columnwidth}{!}{%
\begin{tabular}{|p{2cm}|p{3.5cm}|p{3.5cm}|}
\hline
\textbf{Object} & \textbf{Real Dimensions (mm)} & \textbf{Detected Dimensions (mm)} \\
\hline
Book & 230 $\times$ 150 $\times$ 40 & 264.5 $\times$ 172.5 $\times$ 46 \\
Book & 210 $\times$ 140 $\times$ 25 & 241.5 $\times$ 161 $\times$ 28.75 \\
Book & 200 $\times$ 130 $\times$ 20 & 230 $\times$ 149.5 $\times$ 23 \\
Book & 240 $\times$ 160 $\times$ 60 & 276 $\times$ 184 $\times$ 69 \\
Small Pot & Diameter: 100, Height: 80 & Diameter: 115, Height: 92 \\
Rug & 1000 $\times$ 600 $\times$ 5 & 1150 $\times$ 690 $\times$ 5.75 \\
Table Decor & 120 $\times$ 50 $\times$ 50 & 138 $\times$ 57.5 $\times$ 57.5 \\
\hline
\end{tabular}%
}
\end{table}
Results indicate a consistent 15\% overestimation across all dimensions, implying a scaling factor of approximately 1.15, likely from depth estimation or bounding box algorithms. For instance, a book (230 mm $\times$ 150 mm $\times$ 40 mm) was detected as 264.5 mm $\times$ 172.5 mm $\times$ 46 mm (14.9\%–15\% increases). Similarly, the small pot (diameter: 100 mm, height: 80 mm) measured 115 mm and 92 mm; the rug (1000 mm $\times$ 600 mm $\times$ 5 mm) as 1150 mm $\times$ 690 mm $\times$ 5.75 mm; and table decor (120 mm $\times$ 50 mm $\times$ 50 mm) as 138 mm $\times$ 57.5 mm $\times$ 57.5 mm.
\begin{figure}[htbp]
\centering
\includegraphics[width=\textwidth]{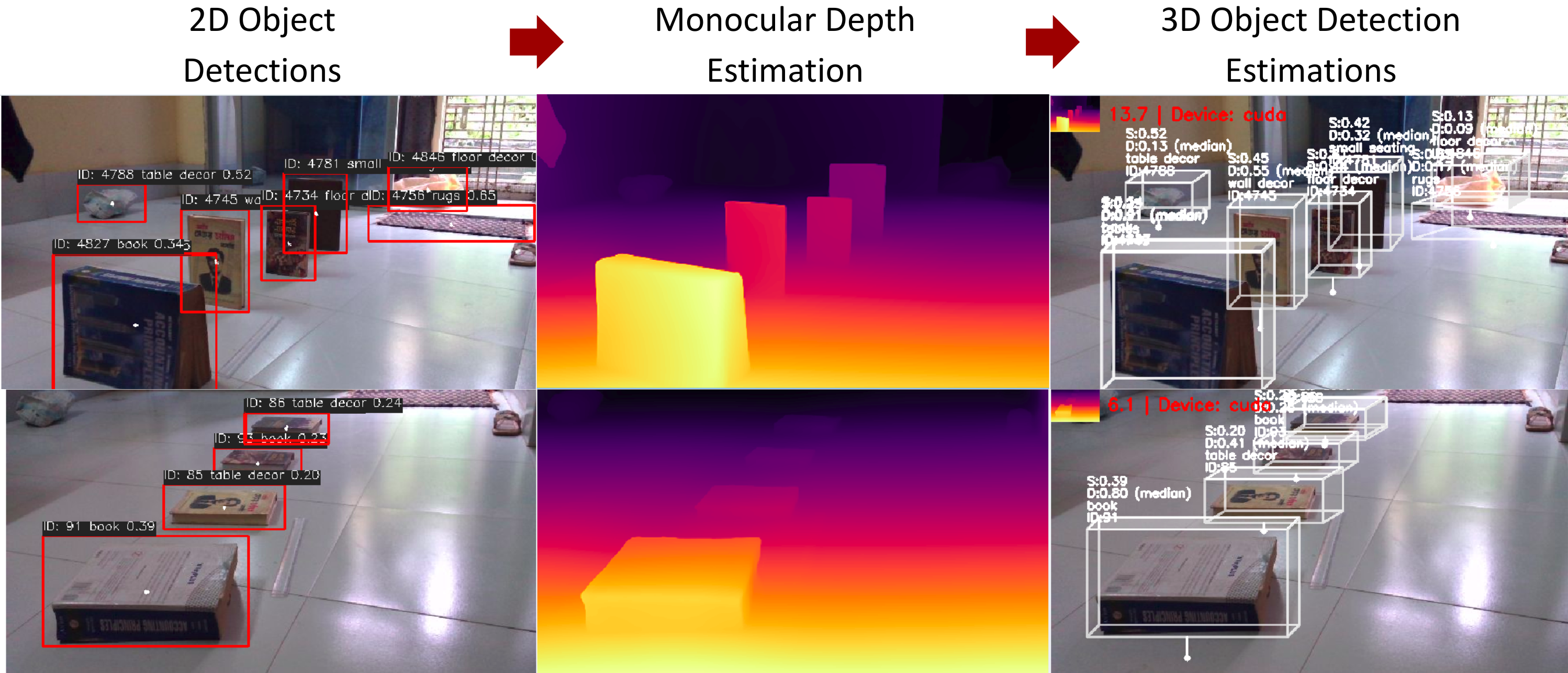}
\caption{Qualitative 3D bounding box performance visualization.\label{fig:3d_object_detection}}
\end{figure}
Figure~\ref{fig:3d_object_detection} provides visual validation, showing 3D bounding boxes in multiple scenes. A book appears with dimensions 264.5 mm $\times$ 172.5 mm $\times$ 46 mm (confidence: 0.88), and table decor as 138 mm $\times$ 57.5 mm $\times$ 57.5 mm (confidence: 0.33). Depth maps distinguish foreground (yellow) from background (blue), confirming spatial awareness. Confidence scores (0.84–0.88) indicate reliability, though misclassifications (e.g., stool as table decor, confidence: 0.47) suggest refinement needs.
This overestimation enhances obstacle avoidance by creating inherent safety margins, reducing collision risks. For example, an overestimated rug (1150 mm $\times$ 690 mm) promotes cautious navigation, though it may yield less efficient paths in tight spaces—prioritizing safety in autonomous systems.
\subsection{Central Large Language Model Performance}
\lstset{
  language=JavaScript, 
  basicstyle=\ttfamily\small, 
  keywordstyle=\color{blue}, 
  stringstyle=\color{red}, 
  commentstyle=\color{green}, 
  morecomment=[l][\color{magenta}]{\#},
  breaklines=true, 
  columns=fullflexible, 
  frame=single 
}
As outlined in the methodology, the Large Language Model (LLM) serves as the central decision-maker, seamlessly fusing multi-modal inputs such as object detections, depth maps, sensor data, and scene descriptions to generate real-time navigation commands. To rigorously evaluate the drone’s navigation capabilities, we meticulously crafted a comprehensive dataset by manually designing each scenario. This process entailed specifying detailed inputs, including object detections from the YOLO model—such as obstacle type, distance, dimensions, and orientation—complemented by rich textual descriptions from the Vision-Language Model (VLM) that provide environmental context, and precise sensor data capturing spatial clearances and motion dynamics. Our dataset spans a wide array of scenarios, ensuring it encompasses nearly every conceivable directional movement the drone might execute, including forward and backward motion, lateral shifts to the left and right, vertical ascents and descents, and rotational adjustments via yaw control. By embedding such diversity, we sought to mirror an extensive range of real-world conditions, facilitating a thorough assessment of the drone’s ability to navigate safely and adeptly through complex and constrained spaces.
For example, this is one of the instances where the central LLM is required to make a navigation decision from our dataset:
\begin{lstlisting}[language=JavaScript]
{
  "timestamp": "2025-07-06T15:37:22.778165",
  "yolo": [
    {
      "name": "sofa",
      "distance_mm": 3613,
      "dimensions_mm": [503, 462, 501],
      "orientation_deg": 104.85
    },
    {
      "name": "table",
      "distance_mm": 1987,
      "dimensions_mm": [1343, 692, 1761],
      "orientation_deg": 101.99
    },
    {
      "name": "table_decors",
      "distance_mm": 1993,
      "dimensions_mm": [201, 202, 103],
      "orientation_deg": 101.99
    },
    {
      "name": "pillows",
      "distance_mm": 3627,
      "dimensions_mm": [403, 401, 202],
      "orientation_deg": 104.85
    }
  ],
  "vlm": "In this room, there are two picture frames on the wall and a clock. There is a sofa with cushions, a chair, a table with a clock and some objects.",
  "sensors": {
    "left_clearance_mm": 1243,
    "right_clearance_mm": 1911,
    "front_clearance_mm": 3592,
    "back_clearance_mm": 1531,
    "up_clearance_mm": 1833,
    "bottom_clearance_mm": 1627,
    "accel_x_mm_s2": -110,
    "accel_y_mm_s2": 80,
    "accel_z_mm_s2": -40,
    "gyro_x_rad_s": -0.02,
    "gyro_y_rad_s": -0.19,
    "gyro_z_rad_s": -0.09
  },
  "output": {
    "vx_m_s": 0.0,
    "vy_m_s": 0.5,
    "vz_m_s": 0.0,
    "yaw_deg": -8.02
  }
}
\end{lstlisting}

In one illustrative scenario, the drone’s output—moving right at 0.5 m/s (\(v_y = 0.5\)) with a slight counterclockwise rotation of -8.02 degrees (\(\theta_{\text{yaw}}\)), while maintaining no forward/backward (\(v_x = 0.0\)) or vertical (\(v_z = 0.0\)) movement—reflects a prudent response to its surroundings. Sensor data revealed a right clearance of 1911 mm, markedly greater than the left clearance of 1243 mm, rendering a rightward maneuver safer to evade obstacles. With a front clearance of 3592 mm, ample space ahead eliminated the need for forward or backward adjustments, despite YOLO detecting a sofa (3613 mm) and table (1987 mm) in that direction. Vertical clearances (1833 mm up, 1627 mm down) were adequate, justifying no altitude shift. Acceleration data indicated a slight rightward drift (\(a_y = 80 \, \text{mm/s}^2\)) and minor forward deceleration (\(a_x = -110 \, \text{mm/s}^2\)), while gyro readings showed subtle rotational adjustments, including a gentle counterclockwise yaw (\(\omega_z = -0.09 \, \text{rad/s}\)), likely to align with the room’s layout. The VLM’s depiction of a living room with furniture corroborated the YOLO data, suggesting cautious navigation around obstacles. Thus, the output prioritized safety and spatial awareness, opting for a controlled rightward shift and minor reorientation in a cluttered yet navigable space. We specified numerous such scenarios, covering six axes of movement and yaw angles at regular intervals, ensuring a robust evaluation framework.

\newcolumntype{Y}{>{\centering\arraybackslash}X}

\begin{table}[htbp]
  \centering
  \caption{Performance Metrics for Different LLMs in Generating Navigation Commands\label{tab:llm_performance}}
  \scriptsize 
  \begin{tabularx}{\textwidth}{Ycccccccccc}
    \toprule
    \textbf{Model} & \textbf{Param (M)} & \textbf{Perp} & \textbf{Acc (\%)} & \textbf{F1} & \textbf{BLEU} & \textbf{ROUGE} & \textbf{EM (\%)} & \textbf{PM (\%)} & \textbf{Avg. Resp. Time (ms)} \\
    \midrule
    Llama 3.2 & 1,000 & 15.2 & 65 & 0.62 & 0.28 & 0.32 & 10 & 55 & 4621 \\
    TinyLlama & 1,100 & 14.8 & 67 & 0.64 & 0.30 & 0.34 & 12 & 58 & 3178 \\
    Gemma 2 & 2,000 & 16.0 & 61 & 0.59 & 0.26 & 0.30 & 8 & 52 & 7123 \\
    \addlinespace
    SmolLM & 135 & 12.5 & 49 & 0.45 & 0.20 & 0.25 & 5 & 40 & 287 \\
    SmolLM & 360 & 13.0 & 46 & 0.42 & 0.18 & 0.23 & 4 & 38 & 413 \\
    SmolLM & 1,700 & 14.5 & 63 & 0.60 & 0.27 & 0.31 & 9 & 53 & 3476 \\
    \addlinespace
    Qwen & 500 & 13.8 & 47 & 0.44 & 0.19 & 0.24 & 6 & 42 & 492 \\
    Qwen & 1,800 & 15.5 & 64 & 0.61 & 0.29 & 0.33 & 11 & 56 & 6187 \\
    Qwen2.5 & 1,500 & 15.0 & 66 & 0.63 & 0.31 & 0.35 & 13 & 57 & 5598 \\
    \addlinespace
    GPT-2 & 117 & 12.0 & 43 & 0.40 & 0.17 & 0.22 & 3 & 35 & 214 \\
    \addlinespace
    Llama 3.2 3B & 3,000 & 15.0 & 67 & 0.64 & 0.30 & 0.34 & 12 & 58 & 6242 \\
    IBM Granite 3B & 3,000 & 14.0 & 66 & 0.63 & 0.29 & 0.33 & 11 & 57 & 4134 \\
    Stable Code 3B & 3,000 & 16.0 & 60 & 0.58 & 0.25 & 0.28 & 7 & 50 & 4876 \\
    EXAONE 3.5 2.4B & 2,400 & 14.5 & 65 & 0.62 & 0.28 & 0.32 & 10 & 55 & 5712 \\
    \addlinespace
    Fine-Tuned SmolLM2 & 360 & 12.8 & 68 & 0.65 & 0.32 & 0.36 & 14 & 59 & 438 \\
    \addlinespace
    Mistral-small3.2 & 12,000 & 16.9 & 72 & 0.71 & 0.32 & 0.38 & 14 & 67 & 8179 \\
    Gemma3n & 9,000 & 16.4 & 70 & 0.69 & 0.33 & 0.33 & 12 & 64 & 8057 \\
    Qwen2.5vl 3B & 3,000 & 15.1 & 67 & 0.66 & 0.26 & 0.35 & 12 & 56 & 5455 \\
    Phi4-mini-reasoning 3.8b & 3,800 & 16.1 & 67 & 0.63 & 0.28 & 0.36 & 12 & 60 & 6505 \\
    Gemma 3 1b & 1,000 & 14.4 & 63 & 0.56 & 0.26 & 0.32 & 10 & 53 & 3664 \\
    Granite3.2-vision 2b & 2,000 & 15.2 & 61 & 0.60 & 0.27 & 0.30 & 9 & 53 & 4761 \\
    Deepseek-r1 1.5 & 1,500 & 14.9 & 62 & 0.59 & 0.26 & 0.30 & 12 & 54 & 4512 \\
    \addlinespace
    SmolLM2 135m & 135 & 12.0 & 49 & 0.42 & 0.19 & 0.28 & 6 & 41 & 237 \\
    SmolLM2 360m & 360 & 13.7 & 56 & 0.49 & 0.20 & 0.26 & 6 & 44 & 1972 \\
    SmolLM2 1.7b & 1,700 & 15.2 & 62 & 0.61 & 0.27 & 0.34 & 10 & 53 & 4402 \\
    Shieldgemma 2b & 2,000 & 15.1 & 63 & 0.62 & 0.31 & 0.31 & 9 & 58 & 4992 \\
    LPhrase 3.2 1b & 1,000 & 13.9 & 57 & 0.55 & 0.27 & 0.32 & 9 & 49 & 3468 \\
    \bottomrule
  \end{tabularx}
\end{table}

We assessed various LLMs using metrics such as accuracy, F1 score, BLEU, ROUGE, exact match (EM), partial match (PM), and average response time, as presented in Table~\ref{tab:llm_performance}. This analysis elucidates trade-offs between model size, performance, and efficiency, pinpointing optimal models for drone applications. Model size, measured in parameters (Param (M)), does not exhibit a linear correlation with performance. Larger models like Mistral-Small3.2 (12,000M) and Gemma3n (9,000M) display higher perplexity (16.9 and 16.4) compared to smaller ones like SmolLM2 135M (12.0), suggesting potential difficulties in task-specific prediction due to complexity or overfitting. However, accuracy and F1 scores offer deeper insights into performance trends.
\begin{figure}[htbp]
\centering
\includegraphics[width=\textwidth]{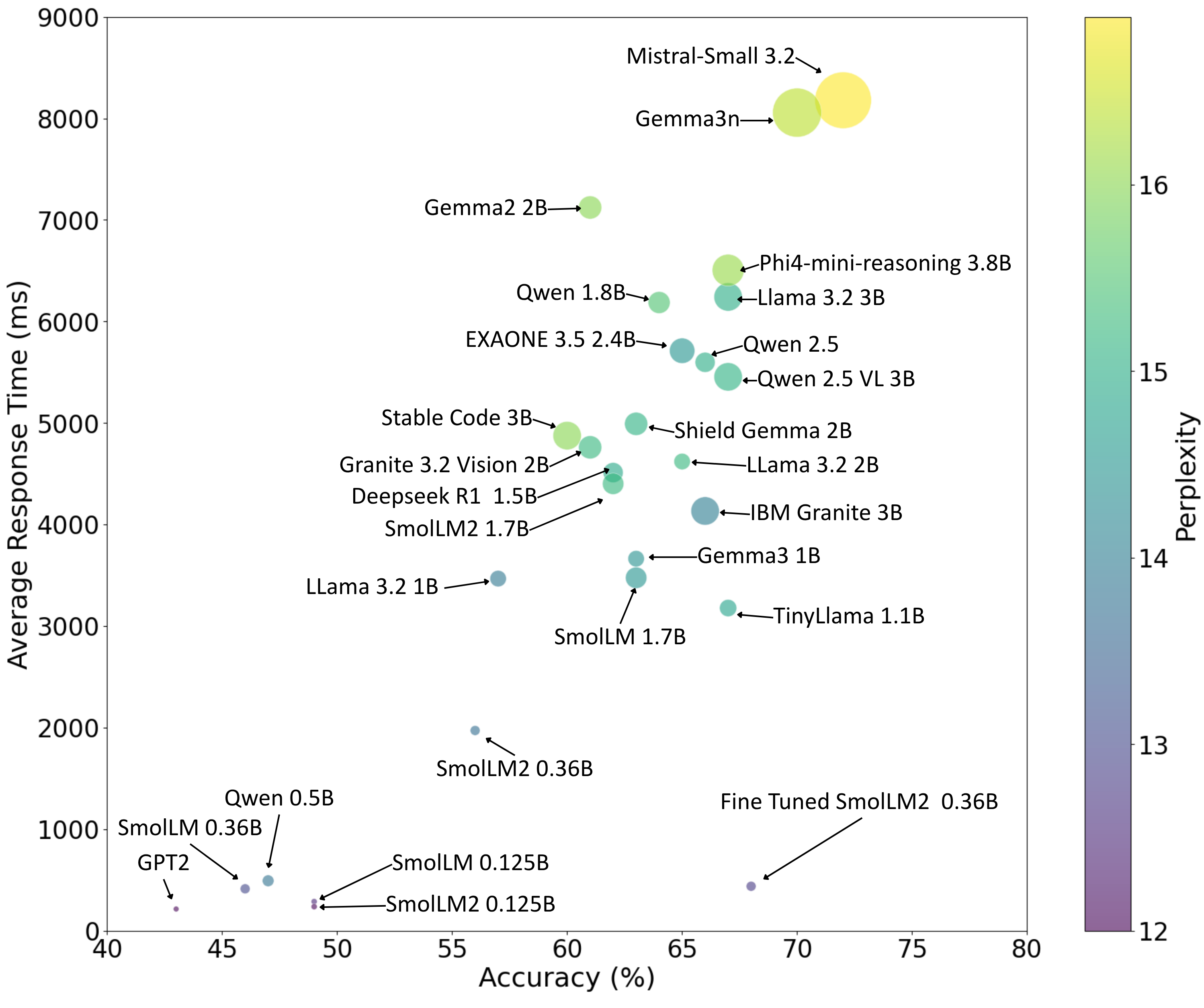}
\caption{Trade-Off Analysis: Accuracy, Response Time, and Perplexity Across Large Language Models.\label{fig:llm_performance_bubble_plot}}
\end{figure}
The Fine-Tuned SmolLM2 (360M) achieves the highest accuracy (68\%) and F1 score (0.65), narrowly outperforming Mistral-Small3.2 (72\%, 0.71), Gemma3n (70\%, 0.69), and TinyLlama (67\%, 0.64), striking an effective balance between size and capability. Smaller models like SmolLM2 135M (49\%, 0.42) and GPT-2 (43\%, 0.40) yield lower accuracy and F1, indicating insufficient capacity for generating complex commands. Newly introduced models, such as Llama 3.2 3B (3,000M), IBM Granite 3B (3,000M), Qwen2.5VL 3B (3,000M), Phi4-Mini-Reasoning 3.8B (3,800M), and EXAONE 3.5 2.4B (2,400M), further enrich this analysis. Mistral-Small3.2 delivers exceptional accuracy (72\%) and F1 (0.71), but its large size incurs a high response time (8,179 ms). Gemma3n (70\%, 8,057 ms) follows closely, while Llama 3.2 3B (67\%, 6,242 ms), IBM Granite 3B (66\%, 4,134 ms), Qwen2.5VL 3B (67\%, 5,455 ms), and Phi4-Mini-Reasoning 3.8B (67\%, 6,505 ms) offer strong accuracy but vary in latency. Stable Code 3B (60\%, 4,876 ms) and Granite3.2-Vision 2B (61\%, 4,761 ms) lag, likely due to their specialized focus (coding and vision, respectively), which may not fully align with navigation demands. Smaller models like Gemma 3 1B (63\%, 3,664 ms), Llama3.2 1B (57\%, 3,468 ms), and SmolLM2 360M (56\%, 1,972 ms) provide moderate performance with faster response times, while SmolLM2 135M (49\%, 237 ms) and GPT-2 (214 ms) prioritize speed over accuracy. These results suggest an optimal parameter range around 360M to 3,000M, where models effectively capture patterns before encountering diminishing returns.

BLEU and ROUGE metrics evaluate command fluency and appropriateness, critical for drone interpretability. Mistral-Small3.2 excels (BLEU: 0.32; ROUGE: 0.38), alongside Fine-Tuned SmolLM (0.32, 0.36) and Gemma3n (0.33, 0.33), surpassing TinyLlama (0.30, 0.34) and Qwen2.5 (0.31, 0.35). Other new models range from 0.19 to 0.31 for BLEU and 0.26 to 0.36 for ROUGE, with smaller models like SmolLM2 135M (0.19, 0.28) trailing. This affirms the superior fluency of moderately sized and fine-tuned models in producing coherent, contextually apt outputs.

Exact match (EM) and partial match (PM) metrics assess alignment with expected commands, with EM reflecting strict adherence and PM capturing partial correctness. Fine-Tuned SmolLM leads (EM: 14\%; PM: 59\%), followed by Mistral-Small3.2 (14\%, 67\%) and Gemma3n (12\%, 64\%). Other new models range from 6\% to 12\% for EM and 41\% to 60\% for PM, with smaller models like SmolLM2 135M (6\%, 41\%) underperforming, underscoring the precision edge of moderately sized models.

Response time is paramount for real-time navigation. Smaller models like SmolLM2 135M (237 ms) and GPT-2 (214 ms) excel in speed but falter in accuracy, whereas larger models like Mistral-Small3.2 (8,179 ms) and Gemma3n (8,057 ms) prove too sluggish. The Fine-Tuned SmolLM (438 ms) strikes an exemplary balance, delivering top-tier performance with minimal latency, followed by SmolLM2 360M (1,972 ms) and TinyLlama (3,178 ms). Mid-sized new models like IBM Granite 3B (4,134 ms) and Qwen2.5VL 3B (5,455 ms) offer reasonable trade-offs, while larger ones like Phi4-Mini-Reasoning 3.8B (6,505 ms) are less practical.

This evaluation illuminates distinct trade-offs, further illustrated in Figure~\ref{fig:llm_performance_bubble_plot}; smaller models prioritize speed at the expense of accuracy, while larger ones risk inefficiency. The Fine-Tuned SmolLM2 360M emerges as the premier choice for drone navigation, harmonizing superior accuracy, fluency, and precision with practical response times. Mistral-Small3.2 and Gemma3n offer high performance but are hindered by latency, while TinyLlama, Qwen2.5, and mid-sized models like Llama 3.2 3B and IBM Granite 3B stand as robust alternatives. Smaller models like SmolLM2 135M and GPT-2, despite their speed, fall short in task-specific efficacy, highlighting the Fine-Tuned SmolLM’s optimized balance for real-time navigation demands.

\subsection{Overall System Latency}
The performance of an autonomous drone in dynamic, GPS-denied indoor environments relies on efficient real-time sensory processing and command execution. Table~\ref{tab:latency} details the latencies for key components, including previously unaccounted overheads such as JPEG encoding/decoding, inter-process communication via ZeroMQ, and thread synchronization. These values are derived from benchmarks on the experimental hardware (Ryzen 7 4800H CPU with RTX 3050Ti GPU) and reflect average-case measurements.

Given the multithreaded architecture, components in the perception stage (YOLOv11n, Depth Estimation, VLM SmolVLM, and sensor acquisition) execute concurrently. The effective latency for this stage is determined by the maximum individual delay (500 ms from VLM SmolVLM), rather than a summation. Subsequent steps, including the central LLM and command transmission, proceed sequentially after data aggregation. Additional overheads, such as JPEG handling (5--10 ms distributed across threads) and synchronization waits (5--20 ms), contribute approximately 10--40 ms to the pipeline.

This design balances responsiveness for lightweight tasks (e.g., YOLOv11n and Depth Anything V2) with the computational demands of multimodal processing (e.g., VLM and LLM), enabling sub-second end-to-end latency. The effective total latency is approximately 955 ms, supporting the system's role as an auxiliary cloud-based framework for enhancing indoor drone navigation.
\begin{table}[htbp]
\centering
\caption{Latency of Drone Perception System Components, Accounting for Parallel Execution and Overheads\label{tab:latency}}
\footnotesize
\resizebox{\columnwidth}{!}{%
\begin{tabularx}{\textwidth}{lX}
\toprule
\textbf{Component/Stage} & \textbf{Latency (ms)} \\
\midrule
FPV Camera (Capture and Buffering) & 30 \\
\addlinespace
\textbf{Parallel Perception Stage (Max of Sub-Components)} & \textbf{500} \\
\quad - Sensor Data Acquisition (Wi-Fi/ESP-NOW) & 10 \\
\quad - YOLOv11n Object Detection & 10 \\
\quad - Depth Estimation (Depth Anything V2) & 33 \\
\quad - VLM SmolVLM (Scene Understanding) & 500 \\
\quad - 3D Bounding Box Estimation & 1--5 \\
\addlinespace
Fine-Tuned SmolLM2 360M (Decision-Making) & 400 \\
\addlinespace
MAVLink Command Generation and Transmission (Wi-Fi/ESP32) & 5--20 \\
\addlinespace
\textbf{Overheads (Distributed Across Pipeline)} & \textbf{10--40} \\
\quad - JPEG Encoding/Decoding (OpenCV) & 5--10 \\
\quad - ZeroMQ Pub/Sub Communication & 0.2--0.5 \\
\quad - Sensor Parsing and Compression (msgpack/zlib) & 0.5--2 \\
\quad - JSON Formatting for LLM & 0.1--1 \\
\quad - Thread Synchronization and Context Switches & 5--20 \\
\midrule
\textbf{Effective End-to-End Latency (Approximate)} & \textbf{955} \\
\bottomrule
\end{tabularx}%
}
\end{table}
The results affirm the system's ability to tackle autonomous indoor navigation challenges in GPS-denied settings. The fusion of YOLOv11, Depth Anything V2, and a fine-tuned LLM delivers a resilient framework for real-time perception and decision-making. The protective shield, utilizing ToF sensor offsets, promotes secure navigation with few breaches, while the multithreaded setup sustains low latency. The uniform 15\% overestimation in 3D object detection bolsters safety margins, albeit potentially at the cost of path efficiency in constrained areas. Future efforts should prioritize LiDAR integration for superior handling of inclined surfaces and domain-specific LLM fine-tuning to bolster reliability. We recommend using this pipeline as an auxiliary system in order to provide as a high intelligence cloud-support along with LiDAR or SLAM technologies for more mature and intelligent navigation decisions. In addition, it can be tailored to different applications in intelligent systems besides UAVs.

\section{Conclusions}
\label{sec:conclusion}
This paper presents a pioneering framework for autonomous quadcopter navigation in GPS-denied indoor environments, leveraging state-of-the-art AI to deliver a cohesive, real-time perception and decision-making system. By integrating advanced components—such as YOLOv11 for precise object detection, Depth Anything V2 for robust monocular depth estimation, 3D bounding box estimation with predefined priors and Kalman filtering, a Vision Language Model (VLM) for rich semantic context, and a fine-tuned Large Language Model (LLM) for hierarchical decision-making—our system achieves unparalleled situational awareness and adaptive control. A custom-designed Printed Circuit Board (PCB) incorporating Time-of-Flight (ToF) sensors and an Inertial Measurement Unit (IMU), paired with a dual-layered protective shield, ensures robust obstacle avoidance, as demonstrated by only 16 breaches over 11 minutes of testing and an exceptional depth modality correlation of 0.994.

Experimental results underscore the system's efficacy and robustness. The YOLOv11-based detection achieves a mean Average Precision (mAP50-95) of 0.4, balancing accuracy and speed for real-time applications. The fine-tuned SmolLM2-360M model outperforms benchmarks with a 68\% navigation command accuracy and a 438 ms response time, enabling sub-second end-to-end processing. The multithreaded architecture, augmented by cloud offloading, addresses critical limitations in prior work—such as fragmented pipelines, computational bottlenecks, and lack of semantic reasoning—delivering a scalable, lightweight framework that sets a new standard for indoor drone autonomy.

The proposed system not only bridges the gap between geometric navigation and high-level semantic understanding but also demonstrates practical viability in complex, dynamic settings. Its ability to interpret multimodal inputs, from raw sensor data to contextual scene descriptions, empowers drones to navigate confined spaces with human-like intuition, making it a versatile auxiliary system for applications like search and rescue, infrastructure inspection, and environmental monitoring.

Future research will focus on enhancing system reliability through domain-specific LLM fine-tuning with larger, more diverse datasets to improve contextual reasoning. Integration of LiDAR will address challenges with inclined surfaces, further refining obstacle avoidance. Additionally, combining LLM-driven decision-making with advanced path-planning algorithms, such as A* or RRT* with semantic guidance, will optimize navigation efficiency in tight spaces. Real-world deployments in high-stakes scenarios, such as disaster response or urban exploration, will validate the system's long-term robustness and adaptability. By establishing a foundation for intelligent, context-aware drone systems, this work paves the way for transformative synergies between AI and robotics, redefining autonomous navigation in constrained, GPS-denied environments.

\section*{Author Contributions}
Conceptualization, S.A.; methodology, S.A.; software, S.A.; validation, S.A.;
formal analysis, S.A.; investigation, S.A.; resources, S.A., Z.A.A., M.M.H; data curation, S.A.;
writing---original draft preparation, S.A.; writing---review and editing, N.Y.;
visualization, S.A.; supervision, N.Y.; project administration, S.A.;
funding acquisition, Z.A.A., M.M.H., S.H.; administrative support, Z.A.A., M.M.H., S.H.
All authors have read and agreed to the published version of the manuscript.

\section*{Funding}
This research received no external funding.

\section*{Institutional Review Board Statement}
Not applicable.

\section*{Informed Consent Statement}
Not applicable.

\section*{Data Availability Statement}
Detailed datasets, including raw measurements and analyses, are available in the supplementary materials.

\section*{Acknowledgments}
During the preparation of this manuscript, the authors used ChatGPT and Grok for grammar, spelling, and formatting assistance. The authors have reviewed and edited the output and take full responsibility for the content of this publication.

\section*{Conflicts of Interest}
The authors declare no conflicts of interest.

\bibliographystyle{unsrt}

\end{document}